%% file: IEEEtran_main.tex
\documentclass[journal]{IEEEtran}
\ifCLASSINFOpdf
\else
\fi
\usepackage{cite}

\input{macros}

\begin{document}
%
\title{Confusing Image Quality Assessment: \\Towards Better Augmented Reality Experience}
%
%
%

\author{Huiyu~Duan,
        Xiongkuo~Min,
        Yucheng~Zhu,
        Guangtao~Zhai,~\IEEEmembership{Senior~Member,~IEEE}, \\
        Xiaokang~Yang,~\IEEEmembership{Fellow,~IEEE}, and 
        Patrick~Le~Callet,~\IEEEmembership{Fellow,~IEEE}

\thanks{Manuscript received October 6, 2021; revised June 2, 2022 and September 12, 2022; accepted October 19, 2022.
This work was supported in part by the National Natural Science Foundation of China under Grant 62225112, Grant 61831015, Grant 62271312, Grant 61901260, and Grant 62101326, in part by the National Key R\&D Program of China 2021YFE0206700, in part by the Shanghai Pujiang Program under Grant 22PJ1407400, and in part by the China Postdoctoral Science Foundation 2022M712090.
The associate editor coordinating the review of this manuscript and approving it for publication was Dr. Chaker Larabi.
\textit{(Corresponding authors: Xiongkuo Min; Guangtao Zhai.)}

Huiyu Duan, Xiongkuo Min, Yucheng Zhu, Guangtao Zhai, and Xiaokang Yang are with the Institute of Image Communication and Network Engineering, Shanghai Jiao Tong University, Shanghai 200240, China (e-mail: huiyuduan@sjtu.edu.cn; minxiongkuo@sjtu.edu.cn; zyc420@sjtu.edu.cn; zhaiguangtao@sjtu.edu.cn; xkyang@sjtu.edu.cn).

Patrick Le Callet is with the Polytech Nantes, Université de Nantes, 44306 Nantes, France (e-mail: patrick.lecallet@univ-nantes.fr).}
}

\input{figures}
\input{tables}

\maketitle

\begin{abstract}
With the development of multimedia technology, Augmented Reality (AR) has become a promising next-generation mobile platform.
The primary value of AR is to promote the fusion of digital contents and real-world environments, however, studies on how this fusion will influence the Quality of Experience (QoE) of these two components are lacking.
To achieve better QoE of AR, whose two layers are influenced by each other, it is important to evaluate its perceptual quality first.
In this paper, we consider AR technology as the \textit{superimposition} of virtual scenes and real scenes, and introduce \textit{visual confusion} as its basic theory.
A more general problem is first proposed, which is evaluating the perceptual quality of superimposed images, \textit{i.e.,} confusing image quality assessment.
A ConFusing Image Quality Assessment (CFIQA) database is established, which includes 600 reference images and 300 distorted images generated by mixing reference images in pairs.
Then a subjective quality perception experiment is conducted towards attaining a better understanding of how humans perceive the confusing images.
Based on the CFIQA database, several benchmark models and a specifically designed CFIQA model are proposed for solving this problem.
Experimental results show that the proposed CFIQA model achieves state-of-the-art performance compared to other benchmark models.
Moreover, an extended ARIQA study is further conducted based on the CFIQA study.
We establish an ARIQA database to better simulate the real AR application scenarios, which contains 20 AR reference images, 20 background (BG) reference images, and 560 distorted images generated from AR and BG references, as well as the correspondingly collected subjective quality ratings.
Three types of full-reference (FR) IQA benchmark variants are designed to study whether we should consider the visual confusion when designing corresponding IQA algorithms.
An ARIQA metric is finally proposed for better evaluating the perceptual quality of AR images.
Experimental results demonstrate the good generalization ability of the CFIQA model and the state-of-the-art performance of the ARIQA model.
The databases, benchmark models, and proposed metrics are available at:
\textcolor{MyMagenta}{https://github.com/DuanHuiyu/ARIQA}.

\end{abstract}


%
\IEEEpeerreviewmaketitle

\vspace{-10pt}
\section{Introduction}
\vspace{-3pt}

\IEEEPARstart{W}ith the evolution of multimedia technology, the next-generation display technologies aim at revolutionizing the way of interactions between users and their surrounding environment rather than limiting to flat panels that are just placed in front of users (\textit{i.e.}, mobile phone, computer, \textit{etc.}) \cite{cakmakci2006head,zhan2020augmented}.
These technologies, including Virtual Reality (VR), Augmented Reality (AR), and Mixed Reality (MR), \textit{etc.}, have been developing rapidly in recent years.
Among them, AR pursues high-quality see-through performance and enriches the real world by superimposing digital contents on it, which is promising to become next-generation mobile platform.
With advanced experience, AR shows great potential in several attractive application scenarios, including but not limited to communication, entertainment, health care, education, engineering design, \textit{etc.}

\FigMotivationI

On account of the complex application scenes, it is important to consider the perceptual Quality of Experience (QoE) of AR, which includes measuring the perceptual quality and better improving the experience of AR.
Lately, some works have been presented to study the quality effects of typical degradations that affect digital contents in AR \cite{guo2016subjective,su2019perceptual,alexiou2018point,zhang2014subjective,zerman2019subjective}.
These studies have performed subjective/objective tests on screen displays showing videos of 3D meshes or point clouds with various distortions.
Moreover, with the development of Head Mounted Displays (HMDs) for AR applications, some studies have considered evaluating the QoE of 3D objects using these devices.
For instance, Alexiou \textit{et al.} \cite{alexiou2017towards} have studied geometry degradations of point clouds and have conducted a subjective quality assessment study in a MR HMD system.
Zhang \textit{et al.} \cite{zhang2018towards} have conducted a study towards the QoE model of AR applications using Microsoft HoloLens, which mainly focused on the perceptual factors related to the usability of AR systems.
Gutierrez \textit{et al.} \cite{gutierrez2020quality} have proposed several guidelines and recommendations for subjective testing of QoE in AR scenarios.
However, all these studies only focus on the degradations of geometry and texture of 3D meshes and point clouds inside AR, \textit{e.g.,} noise, compression \textit{etc.}, their see-through scenes are either blank or simple texture, or even without see-through scenes (opaque images/objects).
The studies discussing the relationship between augmented views and see-through views are lacking.

To address the above issues, in this paper, we consider AR technology as the \textit{superimposition} of digital contents and see-through scenes, and introduce \textit{\textbf{visual confusion}} \cite{woods2010extended,peli2017multiplexing} as its basic theory. 
Fig. \ref{fig:1_visual_conf} demonstrates the concept of the visual confusion in AR.
``B'', ``A'' and ``S'' in Fig. \ref{fig:1_visual_conf} represent the background (BG) view, augmented view and superimposed view, respectively.
If both ``B'' and ``A'' are views with blank/simple textures, there is nothing important in the superimposed view.
If one of ``B'' or ``A'' has a complex texture, but another view has a simple texture, there is also no confusion in the superimposed view.
If both ``B'' and ``A'' have complex textures, visual confusion is introduced.
We assume that without introducing specific distortions that has been widely studied in the current QoE studies, visual confusion itself is a type of distortion, and it significantly influences the AR QoE.
Thus, we argue that it is important to study the assessment of the visual confusion towards better improving the QoE of AR.
Note that it does not mean that no confusion is better than having confusion, since the objective of AR is to promote the fusion between the virtual world and the real world. Instead, the balance between them is more important.

To this end, in this work, we first propose a more general problem, which is evaluating the perceptual quality of visual confusion.
A ConFusing Image Quality Assessment (CFIQA) database is established to make up for the absence of relevant research.
Specifically, we first collect 600 reference images and mix them in pairs, which generates 300 distorted images.
We then design and conduct a comprehensive subjective quality assessment experiment among 17 subjects, which produces 600 mean opinion scores (MOSs), \ie, for each distorted image, two MOSs are obtained for two references respectively.
The distorted and reference images, as well as subjective quality ratings together constitute the ConFusing Image Quality Assessment (CFIQA) database.
Based on this database, several visual characteristics of visual confusion are analyzed and summarized.
Then several benchmark models and a specifically designed attention based deep feature fusion model termed CFIQA are proposed for solving this problem.
Experimental results show that our proposed CFIQA model achieves better performance compared to other benchmark models.

Moreover, considering the field-of-view (FOV) of the AR image and the background image are usually different in real application scenarios, we further establish an ARIQA database for better understanding the perception of visual confusion in real world.
The ARIQA database is comprised of 20 raw AR images, 20 background images, and 560 distorted versions produced from them, each of which is quality-rated by 23 subjects.
Besides the visual confusion distortion as mentioned above, we further introduce three types of distortions: JPEG compression, image scaling and contrast adjustment to AR contents.
Four levels of the visual confusion distortion are applied to mix the AR images and the background images.
Two levels of other types of distortions are applied to the AR contents.
To better simulate the real AR scenarios and control the experimental environment, the ARIQA experiment is conducted in VR environment.
We also design three types of objective benchmark models, which can be differentiated according to the inputs of the classical IQA models, to study whether and how the visual confusion should be considered when designing corresponding IQA metrics.
An ARIQA model designed based on the CFIQA model is then proposed to better evaluate the perceptual quality of AR images.
Experimental results demonstrate that our CFIQA model also achieves good generalization ability on the ARIQA database, and the proposed ARIQA model achieves the best performance compared to other methods.

Overall, the main contributions of this paper are summarized as follows.
(i) We discuss the visual confusion theory of AR and argue that evaluating the visual confusion is one of the most important problem of evaluating the QoE of AR.
(ii) We establish the first ConFusing IQA (CFIQA) database, which can facilitate further objective visual confusion assessment studies.
(iii) To better simulate the real application scenarios, we establish an ARIQA database and conduct a subjective quality assessment experiment in VR environment.
(iv) A CFIQA model and an ARIQA model are proposed for better evaluating the perceptual quality in these two application scenarios.
(v) Two objective model evaluation experiments are conducted on the two databases, respectively, and experimental results demonstrate the effectiveness of our proposed methods.

Our data collection softwares, two databases, benchmark models, as well as objective metrics will be released to facilitate future research. We hope this study will motivate other researchers to 
consider both the see-through view and the augmented view when conducting AR QoE studies.

\vspace{-6pt}
\section{Related Work and Organizations of This Paper \label{sec:II}}

\vspace{-2pt}
\subsection{Augmented Reality and Visual Confusion \label{sec:2.1_AR}}
\vspace{-2pt}
This work mainly concerns head-mounted AR applications rather than mobile phone based AR applications.
Considering rendering methods, there are two main aspects of works in the field of AR visualization, including 2D displaying and 3D rendering.
The superimposition between real-world scenes and virtual contents may cause visual confusion.
Depending on display methods (\ie, superimposition methods), AR technology may produce two kinds of visual effects, including binocular visual confusion and monocular visual confusion.
We discuss the relationship between these aspects and our work as follows.

\textit{\textbf{2D displaying.}} The most basic application of AR is displaying digital contents in a 2D virtual plane \cite{ahn2018real}. 
These digital contents include images, videos, texts, shapes, and even 3D objects in 2D format, \textit{etc.}
To display 2D digital contents, a real world plane is needed to attach the virtual plane.
The real world plane and virtual plane are usually in one same \textit{Vieth–Müller} circle (\textit{a.k.a,} isovergence circle), which may cause visual confusion.
This situation is the main consideration of this paper.

\textit{\textbf{3D rendering.}} Compared to 2D displaying, 3D rendering aims at providing 3D depth cues of virtual objects (note that this depth cue is a little bit different from the depth of the above mentioned virtual plane) \cite{kalkofen2013adaptive}.
Although the 3D depth cues will cause the real world scenes and virtual objects to be located in different \textit{Vieth–Müller} circles, the visual confusion effect still exists, which makes this situation more complex (see Section \ref{sec:VII} for more details).

\textit{\textbf{Visual Confusion.}}
Visual confusion is the perception of two different images superimposed onto the same space \cite{visualconfusion}. The concept of visual confusion comes from ophthalmology, which is usually used to describe the perception of diplopia (\textit{i.e.}, double vision) caused by strabismus \cite{economides2012perception}.
Note that in this paper, we only consider the ``seeing of two or more different views/things in one direction'' \cite{woods2010extended,peli2017multiplexing} as the definition of visual confusion, which should be distinguished from visual illusions \cite{kelley2014animal} and most perceptual distortions.
However, some distortions such as ghosting artifacts can be regraded as visual confusion.
Although the multiplexing superimposition can extend visual perception and has been widely used in prism designs for field expansion \cite{peli2017multiplexing}, some studies have reported that the visual confusion that occurs during field expansion makes users uncomfortable, annoying and disturbing \cite{woods2010extended,apfelbaum2015tunnel}.
Thus, it is significant to study the visual confusion effect for AR QoE assessment.

\textit{\textbf{Binocular visual confusion.}}
The visual confusion caused by two views superimposed binocularly (within two eyes respectively) is \textit{binocular} visual confusion, which may lead to binocular rivalry \cite{blake2002visual,peli2017multiplexing}.
Some previous monocular AR devices, such as Google Glass \cite{googleglass} and VUZIX M400 \cite{vuzix}, were mainly constructed based on binocular visual superimposition to avoid the occlusion produced by AR devices.
However, the binocular rivalry caused by binocular visual confusion may strongly affect the QoE \cite{o2009monocular}.

\textit{\textbf{Monocular visual confusion.}}
The visual confusion caused by two views superimposed monocularly (within one eye) is \textit{monocular} visual confusion \cite{peli2017multiplexing}, which may lead to monocular rivalry \cite{blake2002visual,peli2017multiplexing}.
Since monocular rivalry is much weaker than binocular rivalry \cite{o2009monocular} and it possibly occurs only with extended attention \cite{peli2017multiplexing}, most recent binocular AR technologies were built based on monocular visual superimposition to avoid occlusion, such as Microsoft HoloLens \cite{hololens}, Magic Leap \cite{magicleap}, Epson AR \cite{epson}, \textit{etc.}
However, the QoE of monocular visual confusion still lacks thorough discussion, which is mainly considered in this paper.

\FigClarification

\vspace{-8pt}
\subsection{Image Quality Assessment}
Many IQA methods have been proposed in the past decades \cite{wang2015patch,yu2019predicting}, which can be roughly grouped into three categories, including full reference (FR) IQA, reduced reference (RR) IQA, and no reference (NR) IQA. Considering the possible application scenarios where the digital contents and real-world scenes can be easily obtained, in this paper, we mainly focus on the FR-IQA metric.

\textit{\textbf{Classical IQA index.}}
In terms of FR IQA methods, many classical metrics have been proposed and widely used, including mean squared error (MSE), peak signal-to-noise ratio (PSNR), structural similarity (SSIM) index \cite{wang2004image}, feature similarity (FSIM) index \cite{zhang2011fsim}, \textit{etc.}
Regarding NR IQA indexes, there are also many popular methods such as natural image quality evaluator (NIQE) \cite{mittal2012making}, blind quality assessment based on pseudo-reference image (BPRI) \cite{min2017blind}, and NR free-energy based robust metric (NFERM) \cite{gu2014using}, \textit{etc.}

\textit{\textbf{Learnable IQA.}}
Driven by the rapid development of deep neural networks (DNNs) recently, some learning based IQA algorithms have been proposed.
Kang \textit{et al.} \cite{kang2015simultaneous} proposed to use multi-task convolutional neural networks to evaluate the image quality and use $32 \times 32$ patches for training.
Bosse \textit{et al.} \cite{bosse2017deep} proposed both FR and NR IQA metrics by joint learning of local quality and local weights.
Some studies located specific distortions by using convolutional sparse coding and then evaluated the image quality \cite{yuan2015image}.
Recently, some studies also demonstrated the effectiveness of using pre-trained DNN features in calculating visual similarity \cite{johnson2016perceptual,zhang2018unreasonable}.

\textit{\textbf{AR/VR IQA.}} 
As discussed in the introduction, most previous AR/VR IQA works have studied the degradations of geometry and texture of 3D meshes and point clouds inside AR/VR \cite{alexiou2017towards,zhang2018towards,gutierrez2020quality}.
Unlike AR research, many works on VR IQA have also investigated the quality of omnidirectional images (\textit{a.k.a, equirectangular images}) \cite{duan2017ivqad,duan2018perceptual,sun2019mc360iqa,zhou2021no},
since the format of these images is different with traditional images.
In this paper, we propose that in AR technology, confusing image is its special ``image format", and confusing image quality assessment is equally important with 3D meshes or point clouds quality assessment, since it is not only related to 2D displaying but also associated with 3D rendering effects. Moreover, it significantly influences the QoE of AR.

\vspace{-13pt}
\subsection{Relationship Between CFIQA and ARIQA}
Fig. \ref{fig:2_CFIQA_ARIQA} illustrates the relationship between CFIQA and ARIQA.
CFIQA is a more basic and general problem, which aims at predicting the image quality for each image layer of the superimposition between any two images.
Note that superimposed images/scenes are frequently encountered in the real-world, \eg, rain, haze, reflection, \etc, \cite{zou2020deep,duan2022unified,duan2022develop}, but we mainly focus on the superimposition between two complex images in this paper, since it is directly related to AR applications.
ARIQA is a more specific application scenario of CFIQA, which mainly considers the image quality of the AR layer, since we are more concerned with the saliency or visibility problem of the BG layer \cite{ahn2018real,duan2022saliency}.
This saliency or visibility issue is not something we discuss and address in this paper (see Section \ref{sec:VII} for more details).
Moreover, in this paper, the FOV of two image layers in CFIQA is the same since it is a more general problem and the influence of the superimposition on perceptual quality is unknown, while the FOV of BG scenes is larger than that of AR contents in ARIQA since it is a more realistic situation in AR display \cite{zhan2020augmented,duan2022saliency}.

The rest of the paper is organized as follows.
Section \ref{sec:III} describes the construction procedure of the CFIQA database.
Section \ref{sec:IV} introduces the objective CFIQA models including benchmark models and the proposed CFIQA model.
The experimental validation procedure and results of these objective CFIQA models are given in Section \ref{sec:V}.
Then an extended subjective \& objective ARIQA study is presented in Section \ref{sec:VI}.
Section \ref{sec:VII} concludes the paper and discusses several future issues.

\FigDatasetI

\FigDatasetII

\FigDatasetIII

\FigDatasetIV

\FigDatasetV

\vspace{-12pt}
\section{Subjective ConFusing Image Quality Assessment (CFIQA) \label{sec:III}}
\vspace{-4pt}

\subsection{Confusing Image Collection}
\vspace{-2pt}

To address the problem of subjective confusing IQA data absence, we first build a novel ConFusing Image Quality Assessment (CFIQA) database.
Since this paper is the first work to study confusing IQA, we consider the visual confusion as the only distortion type in this section to study whether and how visual confusion influences the perceptual quality.
We collect 600 images from Pascal VOC dataset \cite{everingham2010pascal} as reference images and split them into two groups.
Then we randomly select two reference images from these two groups and mixed them in pair with a blending parameter $\lambda$ to generate a distorted image.
This can be formulated as:
\vspace{-6pt}
\begin{equation}
    \label{eq:1}
    I_D = \lambda \circ I_{R_1} + (1-\lambda) \circ I_{R_2}~,
    \vspace{-6pt}
\end{equation}
where $I_{R_1}$ is the reference image from the first group, $I_{R_2}$ is the reference image from the second group, $\lambda \in [0,1]$ represents the degradation value of mixing, $I_D$ denotes the generated distorted image.
All reference images are resized to the size of $512 \times 512$ and then superimposed.
A total of 300 distorted images are finally generated.

Obviously, $\lambda$ value near 0 or 1 will cause one image to be unnoticeable while closer to the center (\textit{i.e.,} 0.5) will cause near confusion for both views.
Since visual confusion is the main consideration in this section, it is unreasonable to randomly sample $\lambda$ from $[0,1]$.
In this work, to make the $\lambda$ value to be closer to the center values in range $[0,1]$, we sampled $\lambda$ value from a Beta distribution, \textit{i.e.,} $\lambda \sim \mathbf{Beta}(\alpha,\alpha)$.
The parameter $\alpha$ is set to 5 in this database.
Fig. \ref{fig:2_lambda} demonstrates the distribution of $\lambda$ values used in the CFIQA database.

\vspace{-12pt}
\subsection{Subjective Experiment Methodology \label{sec:subjective_I}}
\vspace{-3pt}
\textit{\textbf{Experiment setup.}} A subjective experiment is conducted on the dataset.
There are several subjective assessment methodologies recommended by the ITU \cite{series2012methodology}, for instance, single-stimulus (SS), double-stimulus impairment scale (DSIS) and paired comparison (PC).
Since the reference images are available, we adopt a paired comparison continuous quality evaluation (PCCQE) strategy to obtain the subjective quality ratings.
As shown in Fig. \ref{fig:3_interface}, for each distorted image, we display its two reference images simultaneously and instruct the subjects to give two opinion scores of the perceptual quality of two layer views (\textit{i.e.,} two reference images) in the distorted image, respectively.
We suggest the subjects to only view the distorted image in the center, and two reference images are just used to determine the quality of which layer is being given.
Two continuous quality rating bars are presented to the subject.
The quality bar is labeled with five Likert adjectives: Bad, Poor, Fair, Good and Excellent, allowing subjects to smoothly drag a slider (initially centered) along the continuous quality bar to select their ratings. 
They are seated at a distance of about 2 feet from the monitor, and this viewing distance is roughly maintained during each session.
All images are shown in their raw sizes, \textit{i.e.,} $512 \times 512$ with random sequence during the experiment.

\textit{\textbf{Testing procedure.}}
As suggested by ITU \cite{series2012methodology}, at least 15 subjects are required to conduct subjective IQA experiment.
A total of 17 subjects are recruited to participate in the study.
Before participating in the test, each subject have read and signed a consent form which explained the human study.
All subjects are determined to have normal or corrected-to-normal vision.
General information about the study is supplied in printed form to the subjects, along with instructions on how to participate in the task.
Each subject then experiences a short training session where 20 confusing images (not included in the formal test) are shown, allowing them to become familiar with the user interface and the visual confusion distortion which may occur.
Moreover, subjects have enough rest time every 10 minutes to avoid fatigue during the experiment.

\vspace{-10pt}
\subsection{Subjective Data Processing and Analysis \label{sec:analysis}}
We follow the suggestions given in \cite{series2012methodology} to conduct the outlier detection and subject rejection.
Specifically, we first calculate the kurtosis score of the raw subjective quality ratings for each image to detect it is a Gaussian case or a non-Gaussian case. Then, for the Gaussian case, the raw score for an image is considered to be an outlier if it is outside 2 standard deviations (stds) about the mean score of that image; for the non-Gaussian case, it is regarded as an outlier if it is outside $\sqrt{20}$ stds about the mean score of that image.
A subject is removed if more than 5\% of his/her evaluations are outliers.
As a result, only 1 subject is rejected, and each image is rated by 16 valid subjects. Among all scores given by the remaining valid subjects, about 2.77\% of the total subjective evaluations are identified as outliers and are subsequently removed.
For the remaining 16 valid subjects, we convert the raw ratings into Z-scores, which are then linearly scaled to the range $[0, 100]$ and averaged over subjects to obtain the final mean opinion scores (MOSs) as follows:
\vspace{-8pt}
\begin{equation}
  z_{ij}=\frac{m_{ij}-\mu_{i}}{\sigma_{i}},\quad z_{ij}'=\frac{100(z_{ij}+3)}{6},
\end{equation}
\vspace{-10pt}
\begin{equation}
  MOS_{j}=\frac{1}{N}\sum_{i=1}^{N}z_{ij}',
\vspace{-6pt}
\end{equation}
where $m_{ij}$ is the raw rating given by the $i$-th subject to the $j$-th image, $\mu_{i}$ is the mean rating given by subject $i$, $\sigma_{i}$ is the standard deviation, and $N$ is the total number of subjects.

Fig. \ref{fig:4_subjective} plots the histograms of MOSs over the entire database as well as within different $\lambda$ value ranges, showing a wide range of perceptual quality scores.
We also plot the Gaussian curve fitting for the histogram.
It can be observed that when $\lambda$ is closer to 0.5, the MOSs tend to be more centered (illustrated by the smaller $\sigma$ value), but still have a wide range of perceptual quality scores.

Based on the constructed database, we further qualitatively analyze the characteristic of the visual confusion.
As shown in Fig. \ref{fig:5_example}, we roughly classify the visual confusion into three categories.
The first category is ``strong confusion'', which means that the mixing of two reference layers will cause strong confusion and may affect the quality rating of the superimposed image (distorted image).
The second category is ``confusion but acceptable'', which represents that the visual confusion caused by the superimposition of two reference layers is acceptable, or uninfluenced, or the perceptual quality is even improved.
The third category is ``suppression'', which denotes that in the superimposed image, one reference layer will suppress another reference layer.
This results in the situation that the perceptual quality of one layer is much better than another layer.
First of all, as a general observation from Fig. \ref{fig:5_example}, we notice that the MOS values (\textit{i.e.,} subjective quality scores) are commonly sorted in descending order as: (1) the activated layer in the ``suppression'' category (\textit{i.e.,} the clearer layer), (2) two image layers in the ``confusion but acceptable'' category, (3) two image layers in the ``strong confusion'' category, and (4) the suppressed layer in the ``suppression'' category (\textit{i.e.,} the fainter layer).
Furthermore, with thorough observation, we notice that the saliency relation between two layers of the superimposed image are important to the perceptual quality, which are demonstrated by several IQA metrics in Section \ref{sec:V}.
From the above observations, we conclude that without introducing other distortions, visual confusion itself can significantly influence the quality of confusing images.

\vspace{-12pt}
\subsection{Controlled Experiment}
\vspace{-3pt}
To further validate the assumption that the superimposition between two images will cause visual confusion and significantly influence the perceptual quality, we also conduct a controlled experiment based on pristine images and compare the MOS results with those of the above subjective CFIQA experiment.
Specifically, a new group of 17 subjects is recruited for the controlled experiments.
These subjects are asked to give their opinion scores for all pristine images of the superimposed images used in the previous subjective CFIQA experiment.
Then the MOSs of these pristine images are calculated correspondingly.
Fig. \ref{fig:5_control} demonstrates the comparison results between this controlled experiment and the above CFIQA experiment.
As shown in Fig. \ref{fig:5_control} (a), the MOS distribution of the controlled experiment is more centered on the right side of this figure compared with that of the CFIQA experiment.
Fig. \ref{fig:5_control} (b) further demonstrates the paired t-test comparison result between the MOSs of the CFIQA experiment and those of the controlled experiment, which illustrates that the MOSs of the controlled experiment are significantly larger ($p<0.0001$).
These experimental results further quantitatively validate our assumption that the visual confusion caused by the superimposition significantly influences the perceptual quality.

\vspace{-8pt}
\section{Objective CFIQA Models \label{sec:IV}}
\vspace{-2pt}

\subsection{Benchmark Models \label{sec:IV:B}}
In terms of the CFIQA database, one distorted superimposed image corresponds to two reference image layers and two subjective ratings.
Thus, we modify state-of-the-art FR-IQA metrics as benchmark models to cope with the CFIQA database as follows.

\textit{\textbf{The simplest metric.}} 
The first intuitive idea is applying the mixing value $\lambda$ as the metric for evaluating visual confusion since it is directly related to the generation of distorted images.
As shown in Eq. (\ref{eq:1}), larger $\lambda$ values make image $I_D$ and image $I_{R_1}$ more similar, and make image $I_D$ and image $I_{R_2}$ more different, while larger $(1-\lambda)$ values make image $I_D$ and image $I_{R_2}$ more similar, and make image $I_D$ and image $I_{R_1}$ more different.
Therefore, for a distorted image $I_D$ generated by two image layers $I_{R_1}$ and $I_{R_2}$ via Eq. (\ref{eq:1}), the simplest FR metric is using $\lambda$ and $(1-\lambda)$ to predict the perceptual qualities of $I_{R_1}$ and $I_{R_2}$ in $I_D$, respectively.

\textit{\textbf{Classical FR-IQA metrics.}} 
We test 16 state-of-the-art classical FR-IQA metrics on the CFIQA database, including MSE, PSNR, NQM \cite{damera2000image}, SSIM \cite{wang2004image}, IFC \cite{sheikh2005information}, VIF \cite{sheikh2006image}, IW-MSE \cite{wang2010information}, IW-PSNR \cite{wang2010information}, IW-SSIM \cite{wang2010information}, FSIM \cite{zhang2011fsim}, GSI \cite{liu2011image}, GMSD \cite{xue2013gradient}, GMSM \cite{xue2013gradient}, PAMSE \cite{xue2013perceptual}, LTG \cite{gu2014efficient}, and VSI \cite{zhang2014vsi}.
Considering that for a distorted image, there are two corresponding reference images, we calculate the FR similarities of this distorted image and the two reference images respectively to obtain two predicted quality scores.
Moreover, since visual attention is important in this task as discussed above, we further select 3 widely used and well-performed metrics, \ie, SSIM, FSIM, and GMSM, and incorporate saliency weights into the quality pooling as new metrics, which are denoted as ``SSIM + saliency'', ``FSIM + saliency'', and ``GMSM + saliency'', respectively.

\FigVGG

\textit{\textbf{Deep feature based IQA metrics.}}
Recently, many works demonstrate the consistence between DNNs and human visual perception \cite{johnson2016perceptual,heusel2017gans,zhang2018unreasonable}.
Therefore, we also consider modifying these DNNs as benchmark models for objective CFIQA.
We first build baseline models with several state-of-the-art DNNs, including SqueezeNet \cite{iandola2016squeezenet}, AlexNet \cite{krizhevsky2012imagenet}, VGG (VGG-16 and VGG-19) \cite{simonyan2014very}, and ResNet (ResNet-18, ResNet-34, and ResNet-50) \cite{he2016deep}, which are denoted as ``Baseline (SqueezeNet)'', ``Baseline (AlexNet)'', ``Baseline (VGG-16)'', \etc.
These baseline models are constructed by averaging the subtracted features of the selected layers (see Section \ref{sec:IV:A}, Para. 2 for more details) between distorted images and reference images as follows:
\vspace{-7pt}
\begin{equation}
    d(l) = \frac{1}{H_{l}W_{l}C_{l}}\sum_{h,w,c}f_d^l,
    \vspace{-7pt}
\end{equation}
where $f_d^l$ is the subtracted feature vector for the selected $l$-th layer (see Eq. (\ref{eq:4}) in Section \ref{sec:IV:A}, Para. 3 for more details), and $H_{l}$, $W_{l}$, $C_{l}$ are the height, width, and number of channels of the $l$-th layer.
Considering the features extracted from the last layer of each ``component'' of these networks may not well reflect the overall performance, in this work, we propose a method to improve the baseline performance based on VGG-16, which is denoted as ``Baseline+ (VGG-16)''.
As illustrated in Fig. \ref{fig:7_vgg_analysis}, for each layer $l$ of the 17 layers (13 convolutional layers and 4 max pooling layers) of VGG-16, we calculate the Spearman Rank-order Correlation Coefficient (SRCC) between $d(l)$ and the MOSs over the whole dataset, which allows us to observe if a particular layer of the model provides more relevant feature maps for CFIQA.
Then the 5 most correlated layers are extracted and averaged to compute the predicted score of the model ``Baseline+ (VGG-16)''.
Furthermore, two widely used deep feature fusion-based metrics (\ie, LPIPS \cite{zhang2018unreasonable} and DISTS \cite{ding2020image}) are also included as benchmark models.

\textit{\textbf{Overall,}} for each benchmark model, we obtain 600 predicted quality scores corresponding to 600 subjective quality ratings (MOSs) for 300 distorted images.
The performance of each model is then calculated between all these predicted quality scores and subjective quality ratings using the criteria discussed in Section \ref{sec:V}.

\FigMethodI

\FigMethodCA

\vspace{-10pt}
\subsection{Attention Based Deep Feature Fusion Method (The Proposed CFIQA Model) \label{sec:IV:A}}
\vspace{-2pt}

As shown in Fig. \ref{fig:5_example}, the classical SSIM index and FSIM index are inconsistent with human perception in some cases (see Section \ref{sec:V} for more quantitative comparison).
From the above analysis, we suppose that the assessment of visual confusion is related to both low-level visual characteristics and high-level visual semantic features, since the visual confusion may disturb semantic information and affect the perceptual quality.
This may cause the failure of the classical metrics, since most of them only consider low-level visual features.
As discussed in \cite{johnson2016perceptual,zhang2018unreasonable}, deep features also demonstrate great effectiveness as perceptual metrics.
Moreover, DNN can extract both low-level and high-level features.
Therefore, in this paper, we propose an attention based deep feature fusion method to measure the visual confusion, which is shown in Fig. \ref{fig:6_model}.

\textit{\textbf{Deep feature extraction.}}
We first employ several state-of-the-art pre-trained DNNs to extract both low-level and high-level features, which include SqueezeNet \cite{iandola2016squeezenet}, AlexNet \cite{krizhevsky2012imagenet}, VGG Net \cite{simonyan2014very}, and ResNet \cite{he2016deep}.
SqueezeNet is an extremely lightweight DNN with comparable performance on classification benchmark.
The features from the first $conv$ layer and subsequent ``$fire$'' modules in SqueezeNet are extracted and used.
We also use a shallow AlexNet network which may more closely match the architecture of the human visual cortex \cite{yamins2016using}, and we extract the features from the $conv1-conv5$ layers in AlexNet.
Furthermore, 5 \textit{conv} layers labeled $conv1\_2$, $conv2\_2$, $conv3\_3/conv3\_4$, $conv4\_3/conv4\_4$, $conv5\_3/conv5\_4$ are extracted from two VGG networks (VGG-16 and VGG-19), respectively.
Finally, considering the effectiveness of ResNet in a large amount of computer-vision (CV) tasks, we also explore the utility of features extracted by ResNet in this task.
Three ResNet architectures including ResNet-18, ResNet-34, as well as ResNet-50 are considered. 
We use the features extracted from the first $conv$ layer and subsequent ``$BasicBlock$'' or ``$Bottleneck$'' modules for all of these architectures.
Overall, for a distorted image $I_D$ and corresponding two reference images $I_{R_1}$, $I_{R_2}$, we extract feature stacks $f_D$, $f_{R_1}$, and $f_{R_2}$ from $L$ layers of a network $\mathcal{F}$, respectively.

\textit{\textbf{Computing feature distance.}}
Then we follow the method in \cite{zhang2018unreasonable} and calculate the feature distance vectors between a distorted image and two reference images by subtracting normalized feature stacks.
This can be expressed as:
\vspace{-3pt}
\begin{equation}
    \label{eq:4}
    f^l_{d_i} = \left \|~ f^l_D - f^l_{R_i}  ~\right \|^2_2~,
    \vspace{-3pt}
\end{equation}
where $l \in [1,L]$ represents the $l$-th layer, $i \in \{1,2\}$ denotes the reference category, $f^l_{d_i}$ is the calculated feature distance vector.

\textit{\textbf{Channel attention for learning feature significance.}}
Since the significance of each channel of the feature distance vector is uncertain for this task, it is important to learn the weights of the channels for each feature distance vector and re-organize them.
We adopt a widely used channel attention \cite{hu2018squeeze} method to learn and re-organize features.
As shown in Fig. \ref{fig:7_modelCA}, after the channel attention, two convolutional layers are followed to reduce channel numbers.
The kernel size of all convolutional layers is set to 1.
Through this manipulation, a distance map stack $m_i$ can be obtained for each $f_{d_i}$, where $i \in \{1,2\}$.

\TabExperimentI

\textit{\textbf{Spatial attention.}}
As discussed in Section \ref{sec:analysis}, high-level visual features such as saliency may influence the perceptual quality of visual confusion.
Since it is hard to optimize spatial attention with a relatively small dataset, in this work, we calculate spatial attention by a saliency prediction method.
A state-of-the-art saliency prediction method \cite{droste2020unified} is used to calculate the spatial attention map $W_i$ for a reference image $I_{R_i}$.
By weighting the distance map $m_i$ with a scaled spatial attention map $W_i$, the distance score for each layer $l$ can be calculated as:
\vspace{-5pt}
\begin{equation}
    d_i^l = \frac{\sum_{h,w}{W_i^l}_{hw} \odot {m_i^l}_{hw}}{\sum_{h,w}{W_i^l}_{hw}}~.
    \vspace{-5pt}
\end{equation}
Then the final quality score can be computed as:
\vspace{-5pt}
\begin{equation}
    s_i = \texttt{Avg}_{_l}(d_i^l)~,
    \vspace{-5pt}
\end{equation}
where $\texttt{Avg}$ is the average operation, and $s_i$ is the predicted quality score.

\textit{\textbf{Loss function.}}
Different with \cite{zhang2018unreasonable}, which aims at two alternative forced choice (2AFC) test, our work focuses on a more general quality assessment task, \textit{i.e.,} the MOS prediction task.
Moreover, different from traditional IQA condition, in our dataset, one distorted image corresponds to two reference images.
Thus, the loss function needs to be carefully designed.
An intuitive loss function is to compare (rank) the perceptual qualities of two layers in the distorted image.
Therefore, a \textit{\textbf{ranking loss}} $\mathcal{L}_R$ is adopted to predict the probability that one layer suppress another layer, which is based on the cross-entropy loss function.
Although the above ranking loss can predict the relative perceptual quality of two layers in a distorted image, the overall qualities of different images across the whole dataset are not normalized and compared.
Therefore, another \textit{\textbf{score regression loss}} $\mathcal{L}_S$ is introduced to regress the probability of the quality being bad or excellent, which is also built based on the cross-entropy loss function.
Two linear layers and a sigmoid layer are used to project the predicted value to the ground-truth space.
The overall loss function can be formulated as:
\vspace{-5pt}
\begin{equation}
    \mathcal{L} = \mathcal{L}_{S_1} + \mathcal{L}_{S_2} + \gamma\mathcal{L}_R,
\vspace{-5pt}
\end{equation}
where the hyperparameter $\gamma$ is empirically set as 2.

\textit{\textbf{Edge feature integration.}} The edges of objects can help identify their categories \cite{liu2019richer}. However, when two images are superimposed together, the intersection of the edges of two image layers may strongly influence the perceptual quality.
Therefore, we further extract the features from an edge detection model \cite{liu2019richer} and concatenate them with the features extracted from one of the aforementioned classification backbones as an enhanced model, which is named CFIQA+.

\FigNewCriterionI

\TabExperimentII

\vspace{-8pt}
\section{Experimental Validation on the CFIQA Database \label{sec:V}}
\vspace{-2pt}

\subsection{Experimental Protocol}
\vspace{-2pt}

\textit{\textbf{Experimental settings.}}
As mentioned in Section \ref{sec:IV:A}, our proposed method needs some samples to train the feature fusion module.
Therefore, the CFIQA database is split into two parts at a ratio of 1:1, \ie, each part has 150 distorted images and corresponding 300 reference images. A two-fold cross-validation experiment is conducted. The two splits of the database are used as the training set and the test set respectively in each cross-validation fold.
During training, the feature extraction network is frozen, and we only train the feature fusion part.
We use Adam as the optimizer \cite{kingma2014adam}.
The network is trained for 100 epoch with a learning rate of 0.0001, and additional 50 epoch with decayed learning rate from 0.0001 to 0.
The batch size is set as 10 during training.

\textit{\textbf{Traditional evaluation metrics.}}
To evaluate the various quality predictors, we use a five-parameter logistic function to fit the quality scores:
\begin{equation}
  Q'=\beta_{1}(\frac{1}{2}-\frac{1}{1+e^{\beta_{2}(Q-\beta_{3})}})+\beta_{4}Q+\beta_{5},
\end{equation}
where $Q$ and $Q'$ are the objective and best-fitting quality, $\beta_{i}(i=1,2,3,4,5)$ are the parameters to be fitted during the evaluation.
Five evaluation metrics are adopted to measure the consistency between the ground-truth subjective ratings and the fitted quality scores, including Spearman Rank-order Correlation Coefficient (SRCC), Kendall Rank-order Correlation Coefficient (KRCC), Pearson Linear Correlation Coefficient (PLCC), Root Mean Square Error (RMSE) and Perceptually Weighted Rank Correlation (PWRC) \cite{wu2018perceptually}.
The evaluation is conducted on the entire database and 2 sub-datasets within different $\lambda$ ranges.
Note that evaluating the performance in the range that the $\lambda$ value is near 0.5 is significant, since it is near the essence of visual confusion and is more common in AR applications, while the poor cases in this situation may strongly influence the QoE.
However, most FR-IQA metrics cannot perform well in this range, which may limit their practicality on CFIQA.

\textit{\textbf{New evaluation methodology.}}
As a complementary, the receiver operating characteristic (ROC) analysis methodology \cite{krasula2016accuracy,krasula2017quality} is also adopted for metric evaluation, which is based on two aspects, \ie, \textit{whether two stimuli are qualitatively different and if they are, which of
them is of higher quality}.
The Fig. 1. in the \textit{supplementary material} illustrates the framework of this evaluation methodology.
We first conduct pair-wise comparison for all possible image pairs, and then classify them into pairs with and without significant quality differences.
Then the ROC analysis is used to determine whether various objective metrics can discriminate images with and without significant differences, termed ``\textit{Different vs. Similar ROC Analysis}''.
Next, the image pairs with significant differences are classified into pairs with positive and negative differences, and the ROC analysis is used to test if various objective metrics can distinguish images with positive and negative differences, termed ``\textit{Better vs. Worse ROC Analysis}''.
The area under the ROC curve (AUC) values of two analysis are mainly reported in this paper, of which the higher values indicate better performance.

\FigARI

\vspace{-9pt}
\subsection{Performance Analysis}
\vspace{-3pt}

\textit{\textbf{Results analysis.}}
First of all, it is important to analyze the performance of all IQA models within different $\lambda$ ranges.
We notice that with $\lambda$ values closer to 0.5 (\textit{i.e.,} the probability of causing strong visual confusion increases), nearly all metrics tend to perform worse.
This indicates that the assessment of strong visual confusion is a difficult task for most models.
As shown in Table \ref{tab:1_all},  $\lambda$ can perform as the metric for confusion evaluation, and even acts better than MSE and PSNR, though the performance is still limited.
Among classical IQA indexes, IW-SSIM and VIF show the top performances, which denotes that visual information weighting possibly helps the assessment of visual confusion.
The improvements of introducing saliency into SSIM, FSIM, as well as GMSM demonstrate the importance of visual attention in visual confusion, which is worth further and deeper research.
Surprisingly, the baseline deep features show good consistence with human perceiving on the entire dataset, though they are not well performed on either of the two sub-datasets.
Unexpectedly, the widely used deep metrics LPIPS \cite{zhang2018unreasonable} and DISTS \cite{ding2020image} perform even worse than the baseline methods, which may indicate that visual confusion is a different type of degradation compared to other distortions.
Finally, our method gets relative better results and different backbone architectures show different optimization trends, which denotes that the feature extraction network is also important.
Future studies on exploring different feature extraction methods are also needed.

Fig. \ref{fig:new_evaluation_cfiqa} illustrates the performance evaluated by the new criteria on the CFIQA database.
First, we observe that the proposed CFIQA model significantly outperforms other state-of-the-art models on \textit{Different vs. Similar Analysis} and \textit{Better vs. Worse Analysis} by a large margin.
Furthermore, we notice the AUC values of the CFIQA metric on the \textit{Better vs. Worse} classification task are higher than the \textit{Different vs. Similar} classification task, which indicates that the \textit{Different vs. Similar} classification is a more hard task and there is still room for improvement in this classification task.

\textit{\textbf{Impact of different components.}}
We further verify the impact of each component in our method.
The analysis is conducted based on the backbone of VGG-16 and the results are shown in Table \ref{tab:2_ablation}.
We first remove all components including channel attention, spatial attention and projection components (two-layer MLP), and only regress the weighting layers for feature fusion.
The results shown in the first row of Table \ref{tab:2_ablation} demonstrate that the performance of this method is similar to the baseline method in Table \ref{tab:1_all}.
Then we compare the impacts of channel attention and spatial attention modules for feature fusion.
It can be observed that the spatial attention module contributes more to the final performance compared to the channel attention module.
Furthermore, we compare the contributions of two loss functions.
It can be observed that ranking loss contributes most for constraining the optimization process.
Though contributing less, the score regression loss still improves the performance.

\vspace{-9pt}
\section{Extension Studies on Augmented Reality Image Quality Assessment (ARIQA)  \label{sec:VI}}
\vspace{-1pt}
In the above Study I and Study II, we have discussed a relatively more basic and more general problem, \ie, visual confusion and its influence on the perceptual QoE.
As aforementioned, visual confusion has significant influence on the QoE of human vision.
However, the situation in the above studies is quite different with the real AR applications, which is mainly attributed to the fact that in actual AR applications, the FOV of the AR contents is usually smaller than the FOV of the real scenes \cite{kruijff2010perceptual}.
Thus, we further conduct another subjective and objective IQA study towards evaluating the perceptual quality of AR contents in real-world AR applications.

\vspace{-10pt}
\subsection{Subjective ARIQA \label{sec:VI-A}}
\vspace{-2pt}

\textit{\textbf{Subjective experiment methodology.}}
An intuitive way to conduct subjective AR experiment is wearing AR devices in various environments and then collecting subjective scores.
However, this way suffers from uncontrollable experimental environments and limited experimental scenarios \cite{gutierrez2020quality}, \eg, the head movement may cause different collected background images for different users, and it is hard to introduce various background scenarios in lab environment.
Therefore, we adopt the method of conducting subjective AR-IQA studies in VR environment for controllable experimental environments and diverse experimental scenarios.

Fig. \ref{fig:7_ar_simulation} illustrates the methodology of the subjective experiment in this ARIQA study.
First of all, 20 omnidirectional images are collected as the background scenes including 10 indoor scenarios and 10 outdoor scenarios.
Considering the real applications of AR, we further collect 20 images as the reference AR contents, which include 8 web page images, 8 natural images, and 4 graphic images.
The resolution of all raw AR images is $1440 \times 900$. 
We generate a much larger set of distorted AR contents by applying quality degradation processes that may occur in AR applications.
Three distortion types including image compression, image scaling, image contrast adjustment, are introduced as follows.
(i) \textit{JPEG compression} is a widely used method in image compression, and have been introduced into many quality assessment databases \cite{yang2015perceptual}.
We set the quality level of the JPEG compression at the two levels with quality parameters 7 and 3.
(ii) \textit{Image scaling} is widely used in modern video streaming systems, where videos are often spatially downsampled prior to transmission, and then upscaled prior to display \cite{min2020study}.
Such image scaling can also simulate the distortions introduced by various resolutions of AR devices.
We create distorted images by downsclaing original images to 1/5 and 1/10 of the original resolution, then spatially upscaling them back to the original resolution.
(iii) \textit{Image contrast adjustment} is also an important factor affecting the human visual perception and has been commonly introduced into natural IQA \cite{gu2015analysis} and screen content IQA \cite{yang2015perceptual}.
We also use the gamma transfer function \cite{gu2015analysis} to adjust the contrast, which is defined as $y=[x \cdot 255^{((1/n)-1)}]^n$, where $n=[1/4,4]$ ($n<1$ is negative gamma transfer, $n>1$ is positive gamma transfer).
Hence, for each AR image, we generate 6 degraded images.

\FigARInterface

\FigARII

\FigARIII

\FigARIV

Since the visual confusion strongly influences the human visual perception as aforementioned, the superimposition degradation is also introduced.
We design a program using Unity \cite{Unity} to perform the experimental procedure, including stimuli display, data collection, \etc.
Fig. \ref{fig:7_ar_interface} demonstrates the user interface of the subjective ARIQA experiment.
We first randomly match the 20 AR images and 20 omnidirectional images in pairs to generate 20 scenarios.
Hence, for each omnidirectional image, we have 7 AR images superimposed on it (1 reference image + 6 distorted images).
During the experiment, the perceptual viewport can be formulated as: 
\vspace{-5pt}
\begin{equation}
    I_S = \lambda \circ I_A + (1-\lambda) \circ I_O~,
    \vspace{-5pt}
\end{equation}
where $I_S$ denotes the perceptual viewport, \ie, the superimposed image, $I_A$ represents the AR image, $I_O$ indicates the omnidirectional image, and $\lambda \in [0.26, 0.42, 0.58, 0.74]$ denotes the mixing value used in the experiment, \ie, four superimposed levels are introduced in this subjective experiment.
Overall, 560 experimental stimuli are generated for conducting the subjective experiment ($20~scenarios \times 7~levels \times 4~mixing~ values$).
As demonstrated in Fig. \ref{fig:7_ar_simulation} (a), the omnidirectional image is displayed in 360 degrees as the background scenarios, the AR image is superimposed on the omnidirectional image which is perceived by the perceptual viewport.
Fig. \ref{fig:7_ar_simulation} (b), (c) and (d) show the examples of the omnidirectional images, the AR images, and the perceptual viewport images, respectively.

A total of 23 subjects participate in the experiment, who are not included in the aforementioned CFIQA study.
All subjects are recruited through standard procedures similar to that described in Section \ref{sec:subjective_I}.
Before the formal test, each subject experiences a short training session where 28 stimuli are shown.
The same distortion generation procedure is introduced for the training stimuli as for the test stimuli, and these training stimuli are not included in the test session.
Since the experiment is conducted under VR-HMD environment, the single-stimulus (SS) strategy is adopted to collect the subjective quality ratings of AR images.
A 10-point numerical categorical rating method is used to facilitate the subjective rating in HMD \cite{duan2018perceptual}.
We use HTC VIVE Pro Eye \cite{HTC} as the HMD on account of its excellent graphics display technology and high precision tracking ability.
During the formal test, all 560 experimental stimuli are displayed in a random order for each subject.

\textit{\textbf{Subjective data processing and analysis.}}
Similar to the procedure in Section \ref{sec:analysis}, we first process the collected subjective scores to obtain the MOSs.
Only 1 subject is rejected, and each image is rated by 22 valid subjects.
Among all scores given by the remaining valid subjects, about 3.21\% of the total subjective evaluations are identified as outliers and removed.
Fig. \ref{fig:8_ariqa_dataset} plots the histogram of MOSs over the entire ARIQA database, showing a wide range of perceptual quality scores.

We further analyze the distribution of MOS values across different mixing values and various distortions.
Fig. \ref{fig:9_ariqa_dataset_analysis1} shows the MOS distribution of the images with the degradations of JPEG compression and image scaling under different mixing values.
We notice that as the $\lambda$ value increases, the MOS value also shows an overall upward trend, of which the reason is apparent since larger $\lambda$ value means clearer AR content.
Specifically, for the superimposition of raw AR images and background images, we find that graphic images provide better QoE than web page images, and web page images provide better QoE than natural images in general.
It may reveal that relatively simple AR contents can provide better QoE than complex AR contents.
Moreover, for the superimposed AR images with JPEG compression and scaling, we notice that when the mixing value $\lambda$ is relatively smaller, the MOSs of these images are closer to that of superimposed raw images, though the overall MOSs are smaller than that of the larger $\lambda$ values.
It may reveal that the superimposition degradation is a more influential quality factor compared to other distortions when the $\lambda$ value is relatively small.
However, it also means that the superimposition degradation can hide other distortions.
Fig. \ref{fig:10_ariqa_dataset_analysis2} plots several examples of the MOS values of raw images and contrast adjusted images superimposed on the omnidirectional backgrounds with different mixing values, which shows that appropriate contrast adjustment may even improve the perceptual quality of AR contents.

\FigMethodII

\TabExperimentIII

\TabExperimentIV

\vspace{-12pt}
\subsection{Objective ARIQA Models}
\vspace{-4pt}

\subsubsection{Benchmark models}

As discussed in Section \ref{sec:III} and Section \ref{sec:VI-A}, the visual confusion may affect the human visual perception and may degrade the QoE of AR.
However, whether the IQA metrics should consider the superimposed image, the AR image, and the background image together still needs to be discussed.
Therefore, three benchmark variants are introduced for objective ARIQA.
We assume the background image, the AR image, as well as the mixing value are known, which can be acquired in real applications, and the superimposed image can be correspondingly calculated.
Let $I_{A_D}$ denotes the AR image with distortions, $I_{A_R}$ denotes the raw reference AR image, $I_B$ indicates the background image, $\lambda$ represents the mixing value, hence, the displayed AR image $I_A$ and the perceptual viewport image (superimposed image) $I_S$ can be correspondingly expressed as: $I_A = \lambda \cdot I_{A_D}$, and $I_S = \lambda \cdot I_{A_D} + (1-\lambda) \cdot I_B$, respectively.
Then, three FR-IQA benchmark variants used to calculate AR image quality are defined as: Type I, the similarity between the displayed AR image $I_A$ and the reference AR image $I_{A_R}$; Type II, the similarity between the perceptual viewport image $I_S$ and the reference AR image $I_{A_R}$; Type III, the SVR fusion \cite{chang2011libsvm} of the similarity between the perceptual viewport image $I_S$ and the reference AR image $I_{A_R}$, and the similarity between the perceptual viewport image $I_S$ and the background image $I_B$.
These three variant types can be expressed as:
\vspace{-4pt}
\begin{equation}
    Q_\text{Type I} = \text{FR}(I_A,I_{A_R}),
\end{equation}
\vspace{-16pt}
\begin{equation}
    Q_\text{Type II} = \text{FR}(I_S,I_{A_R}),
\end{equation}
\vspace{-16pt}
\begin{equation}
    Q_\text{Type III} = \text{SVR}(\text{FR}(I_S,I_{A_R}), \text{FR}(I_S,I_B)),
\vspace{-4pt}
\end{equation}
where $Q_\text{Type I}$, $Q_\text{Type II}$, and $Q_\text{Type III}$ denote the quality predictions of the three variants, FR represents the used FR-IQA metric, SVR indicates the support vector regression deployment.

\subsubsection{The proposed ARIQA model}

Intuitively, using superimposed images as the perceptual distorted images, and considering their similarity with both AR references and background references may be more effective for evaluating the perceptual quality of the AR layers.
Hence, as demonstrated in Figure \ref{fig:11_modelII}, for better evaluating the perceptual quality of AR contents, we further improve the learning strategy of the CFIQA model to the ARIQA model by comparing the quality of two homologous superimposed images.
Different from CFIQA, the goal for the ARIQA is to predict the perceptual quality of AR contents rather than both two views, therefore, the two output results of CFIQA are fused to predict the AR image quality.
Considering the effectiveness of the training objectives of the LPIPS \cite{zhang2018unreasonable} and our CFIQA, during the training process, two pathways are introduced to ARIQA for comparing the perceptual quality of two different distorted images of one AR and background reference pair.
Furthermore, we also improve the ARIQA model to the ARIQA+ model by incorporating the features from the edge detection model RCF net \cite{liu2019richer}, which is similar to the way of the aforementioned CFIQA+.

\FigNewCriterionII

\vspace{-10pt}
\subsection{Experimental Validation on the ARIQA Database}

\subsubsection{Experimental settings}
In terms of our ARIQA database, the background image $I_B$ (\ie, viewport of the omnidirectional image $I_O$) and the superimposed image $I_S$ are captured in Unity, then the results of our proposed ARIQA model and benchmark models can be calculated accordingly as aforementioned.

\textit{Benchmark experiment setting.} Besides the state-of-the-art FR-IQA metrics mentioned in Section \ref{sec:IV:B}, we also test the generalization ability of the CFIQA model trained on the CFIQA database on the ARIQA database.
For the SVR experiment (\ie, Type III mentioned above), we conduct a 100-fold cross validation experiment.
For each fold, the ARIQA database is randomly split into a training dataset and a test dataset at a ratio of 4:1.
The final results are calculated by averaging the test results of all 100 cross-validations.

\textit{Deep learning-based experiment setting.} We conduct a five-fold cross-validation experiment for the proposed ARIQA model on the ARIQA database.
For each fold, we split the 560 samples into 280 training samples and 280 testing samples without scene repeating, \ie, 280 training samples and 280 testing samples corresponding to different 10 AR/BG pairs, respectively.
For fair comparison, we also re-train the LPIPS and CFIQA models only using AR image as the reference image, which is similar to the concept of Type II described above.
Note that the CFIQA model here is a modified version, since the original CFIQA aims to compare the similarity between two reference images for one superimposed image, while here we focus on comparing two superimposed images for one AR reference, which is more similar to the concept of LPIPS.

\subsubsection{Performance analysis}

Table \ref{tab:3_AR_base} presents the performance of three types of benchmark variants derived from the state-of-the-art FR-IQA models on the ARIQA database.
Comparing Type I and Type II, we notice that for most FR-IQA metrics, using superimposed images as distorted images can improve the performance of the algorithm.
In addition, as shown in the comparison between Type III and Type I, when superimposed images, AR images, as well as background images are jointly considered, the performance of almost all FR-IQA metrics can be further improved.
Moreover, it can be observed that our proposed CFIQA models trained on the CFIQA database also achieve state-of-the-art performance on the ARIQA database.
This demonstrates the good generalization ability of our proposed CFIQA models and illustrates that visual confusion is one of the most important factors affecting the perceptual quality of AR images.

Table \ref{tab:4_ARIQA} shows the averaged performance of these four models after five-fold cross validation.
It can be observed that the ARIQA model achieves better performance than the LPIPS model and the CFIQA model, and the ARIQA+ achieves the best performance compared to other models.
Fig. \ref{fig:new_evaluation_ariqa} illustrates the performance evaluated by the new criteria on the ARIQA database.
We notice that the proposed ARIQA model significantly outperforms other state-of-the-art models on \textit{Different vs. Similar Analysis} and \textit{Better vs. Worse Analysis} by a large margin.

\vspace{-6pt}
\section{Conclusion and Future Works \label{sec:VII}}
\vspace{-2pt}

In this paper, we discuss several AR devices and applications (see Section \ref{sec:2.1_AR}), and clarify the essential theory underlying AR, \textit{i.e.,} visual confusion.
A more general problem underlying AR QoE assessment is first proposed, which is evaluating the perceptual quality of superimposed images, \ie, confusing image quality assessment.
To this end, we build a confusing image quality assessment (CFIQA) database, and conduct subjective and objective image quality assessment studies based on it.
A CFIQA model is also proposed for better evaluating the perceptual quality of visual confusion.
The results show that without extra degradation, the visual confusion itself can significantly influence the perceptual quality of the superimposed images, and state-of-the-art FR-IQA metrics are not well performed on this problem, especially when the mixing value is closer to 0.5.
Moreover, the proposed CFIQA model performs better on this task.
Next, in order to better study the influence of visual confusion on the perceptual quality of AR images in real application scenarios, an ARIQA study is further conducted.
An ARIQA database is constructed, and three benchmark model variants as well as a specifically designed ARIQA model are proposed.
The results show that it is beneficial to consider visual confusion when designing IQA models for AR, and our proposed ARIQA model achieves better performance compared to other benchmark models.
We hope this work can help other researchers have a better understanding of the visual confusion effect underlying AR technology.
There are many issues related to visual confusion or AR QoE assessment that need to be explored in the future.
Several key aspects are discussed as follows.

\textit{\textbf{More complex degradations and diverse digital contents.}}
In this paper, on the basis of visual confusion degradation, we further incorporate three other distortion types.
However, we may encounter more complex degradations \cite{ponomarenko2015image, hosu2020koniq} in daily life due to the limitation of photographic apparatus, compression processing, transmission bandwidth, display devices, \textit{etc.}
The perceptual peculiarity of these more complex distortions may be different from their original characteristics when mixed with the visual confusion degradation.
Moreover, besides 2D AR contents, 3D rendering may also produce visual confusion effect when the 3D virtual content is not fitting in the real-world environment.
The perceptual quality of these more complex degradations and diverse digital contents under the visual confusion condition needs to be further studied.

\textit{\textbf{More realistic simulation for real-world scenes and AR devices.}}
It should be noted that the simulation used in our ARIQA study can also be improved.
First of all, in the ARIQA study, we use omnidirectional images as background scenes, which cannot simulate real-world depth cues.
It is significant to consider using omnidirectional stereoscopic images or virtual 3D scenes as background scenes to study the 3D visual confusion problem in future studies.
However, the stereoscopic factor is hard to control, which should be carefully designed.
Moreover, some AR devices may be equipped with dynamic dimming functions, such as Microsoft HoloLens \cite{hololens} and Magic Leap \cite{magicleap} \etc.
It is also important to discuss this aspect when studying visual confusion effect for specific devices in future works.
Finally, in real AR applications, the dynamic range of background scenes and AR contents may be different, since the background scenes are usually optically see-through and directly observed by users, while the dynamic range of AR contents is limited by the display.
However, due to the limitation of the VR display, we cannot reproduce the dynamic range of real-wold scenes.
Future works can also consider using real AR devices to study the influence of visual confusion on the QoE of AR.

\textit{\textbf{Visual attention/saliency of visual confusion.}}
As can be observed in Section \ref{sec:V} above, 
incorporating saliency as spatial attention into IQA metrics introduces significant performance improvement.
However, in this work, we just apply the saliency map predicted by current methods on the reference images.
We notice that state-of-the-art saliency prediction models fail to predict the saliency on the superimposed distorted images.
Since the superimposition may change the original texture, the visual attention may be altered correspondingly.
Future studies on predicting the saliency of confusing (superimposed) images are also needed \cite{duan2022saliency}.
Moreover, the relationship between saliency prediction and quality assessment under the visual confusion situation also needs more discussion.
In our ARIQA study, we notice that clearer AR contents bring better perceptual quality for them.
However, it does not mean that we should make the AR as clear as possible, since it may completely occlude the background view and may cause trouble or even danger.
For see-through views, besides the perceptual quality metric, we suppose that the visibility metric may be a good way for evaluating it, which needs more research on it.

\textit{\textbf{How to improve the QoE of AR.}}
Based on the above discussions, we suppose that it is important to consider how to \textit{carefully design} and \textit{harmoniously display} the digital contents of AR to make the QoE of both the virtual world and the real world better, especially for different application scenarios and user requirements.
We discuss some factors and present several recommendations that may be deserved to be studied to improve the QoE of AR in the future as follows.
\textbf{(i)} FOV. 
The main difference between the CFIQA database and the ARIQA database is that the FOVs of two superimposed views in the ARIQA database are different.
We notice that the superimposition of two views with different fields may help distinguish each other.
Future works on how to appropriately design the FOV for specific applications may be helpful.
\textbf{(ii)} 3D depth cues. The depth difference between AR contents and BG scenes can help distinguish two layers and may improve the perceptual quality of them. However, it should be noted that the depth difference may increase the risk of inattentional blindness \cite{mack2003inattentional,wang2021inattentional}.
Thus it may be better to study the influence of this factor in both the quality problem (as discussed in this paper) and the saliency problem \cite{duan2022saliency} together to give a trade-off solution.
\textbf{(iii)} Similarity/correlation between AR contents and BG scenes.
As shown in Fig. \ref{fig:1_visual_conf}, if the background view or augmented view is blank, there is no visual confusion.
However, this situation is unlikely to happen in real cases.
A more reasonable solution is calculating the similarity/correlation between AR view and BG view, then looking for appropriate space to display AR contents or adjusting the brightness/contrast/color of AR contents \cite{ahn2018real}.
Our proposed method can be used in this situation.
\textbf{(iv)} Displaying AR contents considering scenarios. It should be noted that different scenarios (\textit{e.g.,} moving, talking, relaxing, \textit{etc.}) may need different display solutions.
This is a human-centric problem, which may need to combine other computer vision techniques from egocentric problems \cite{grauman2021ego4d} for AR displaying.

\textit{\textbf{Other computer vision tasks and applications.}}
The directly related CV tasks to this work are blind source separation (BSS) \cite{hyvarinen2000independent,duan2022develop} and its sub-tasks, such as image reflection removal \cite{duan2022develop} \textit{etc.}
BSS aims at separating source signals/images from a set of mixed ones, which has been widely explored recently \cite{zou2020deep,duan2022develop}.
Research on FR or NR IQA metrics of visual confusion may contribute to the evaluation of these problems.
Moreover, besides the AR, the visual confusion may also appear in other display technologies, such as projector and transparent display screens \cite{luxlabs}.
Future work on NR-IQA metrics of visual confusion may help assess the QoE of these devices.



\ifCLASSOPTIONcaptionsoff
  \newpage
\fi



%

\bibliographystyle{IEEEtran}
\bibliography{IEEEtran}

%


\vspace{-40pt}
\begin{IEEEbiography}[{\includegraphics[width=1in,height=1.25in,clip,keepaspectratio]{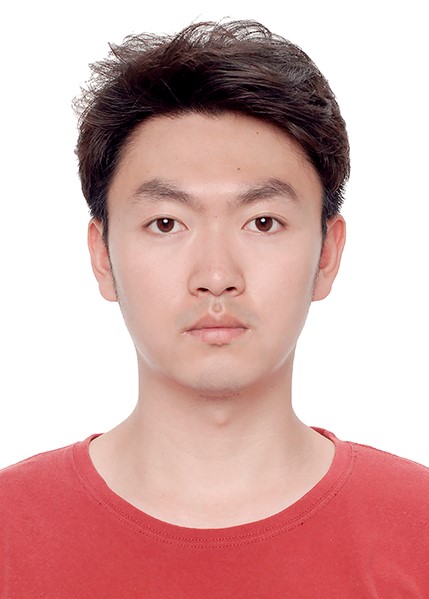}}]{Huiyu Duan}
received the B.E. degree from the University of Electronic Science and Technology of China, Chengdu, China, in 2017. He is currently pursuing the Ph.D. degree with the Department of Electronic Engineering, Shanghai Jiao Tong University, Shanghai, China. From Sept. 2019 to Sept. 2020, he was a visiting Ph.D. student at the Schepens Eye Research Institute, Harvard Medical School, Boston, USA. His research interests include perceptual quality assessment and extended reality (XR).

\end{IEEEbiography}
\vspace{-40pt}

\begin{IEEEbiography}[{\includegraphics[width=1in,height=1.25in,clip,keepaspectratio]{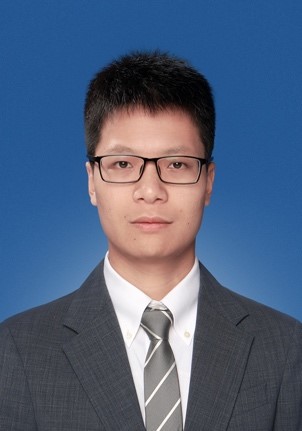}}]{Xiongkuo Min}
received the B.E. degree from Wuhan University, Wuhan, China, in 2013, and the Ph.D. degree from Shanghai Jiao Tong University, Shanghai, China, in 2018, where he is currently a tenure-track Associate Professor with the Institute of Image Communication and Network Engineering. 
His research interests include image/video/audio quality assessment, quality of experience, visual attention modeling, extended reality, and multimodal signal processing.

\end{IEEEbiography}
\vspace{-40pt}

\begin{IEEEbiography}[{\includegraphics[width=1in,height=1.25in,clip,keepaspectratio]{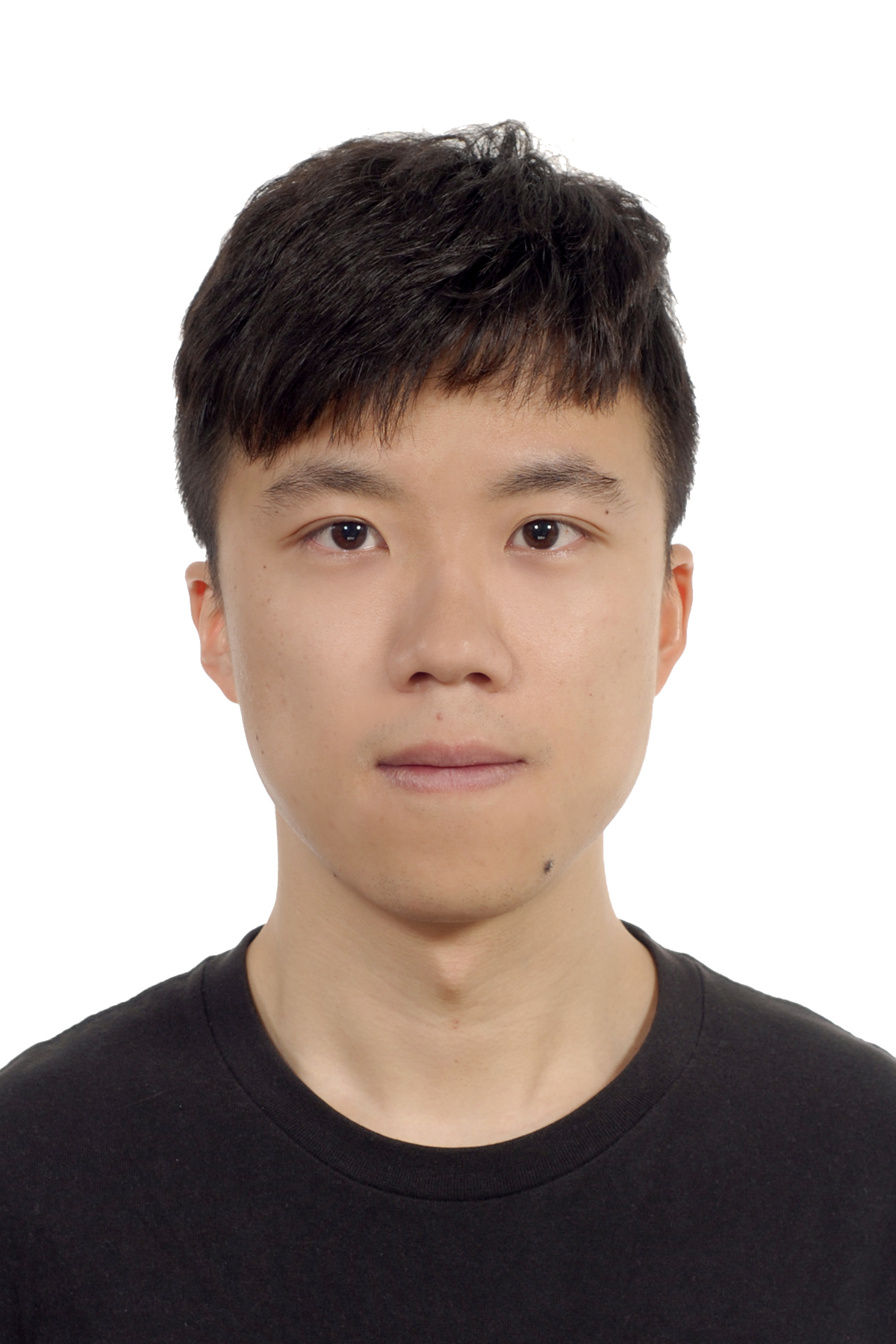}}]{Yucheng Zhu}
received the B.E. degree from the Shanghai Jiao Tong University, Shanghai, China, in 2015, and the Ph.D. degree from Shanghai Jiao Tong University, Shanghai, China, in 2021. He is currently a Post-Doctoral Fellow with Shanghai Jiao Tong University. His research interests include visual quality assessment, visual attention modeling and perceptual signal processing.

\end{IEEEbiography}
\vspace{-40pt}

\begin{IEEEbiography}[{\includegraphics[width=1in,height=1.25in,clip,keepaspectratio]{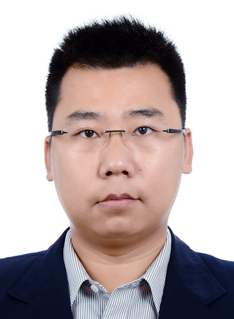}}]{Guangtao Zhai}
(SM’19) received the B.E. and M.E. degrees from Shandong University, Shandong, China, in 2001 and 2004, respectively, and the Ph.D. degree from Shanghai Jiao Tong University, Shanghai, China, in 2009, where he is currently a Research Professor with the Institute of Image Communication and Information Processing. 
He received the Award of National Excellent Ph.D. Thesis from the Ministry of Education of China in 2012. His research interests include multimedia signal processing and perceptual signal processing.

\end{IEEEbiography}
\vspace{-40pt}

\begin{IEEEbiography}[{\includegraphics[width=1in,height=1.25in,clip,keepaspectratio]{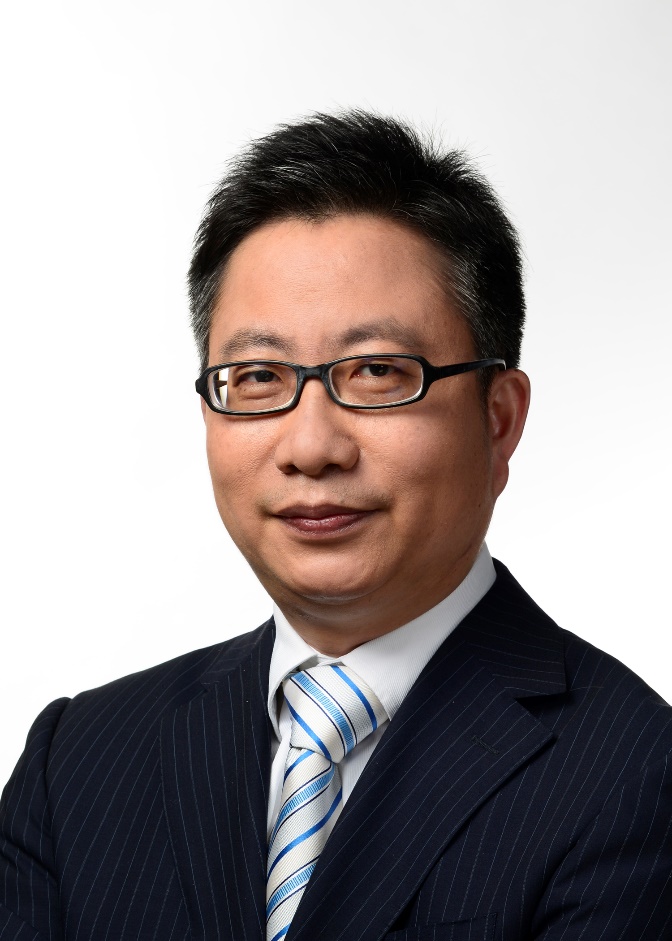}}]{Xiaokang Yang}
(M’00-SM’04-F’19) received the B.S. degree from Xiamen University, Xiamen, China, in 1994, the M.S. degree from the Chinese Academy of Sciences, Shanghai, China, in 1997, and the Ph.D. degree from Shanghai Jiao Tong University, Shanghai, in 2000. 
His current research interests include image processing and communication, computer vision, and machine learning. He is an Associate Editor of the IEEE TRANSACTIONS ON MULTIMEDIA and a Senior Associate Editor of the IEEE SIGNAL PROCESSING LETTERS.

\end{IEEEbiography}
\vspace{-40pt}

\begin{IEEEbiography}[{\includegraphics[width=1in,height=1.25in,clip,keepaspectratio]{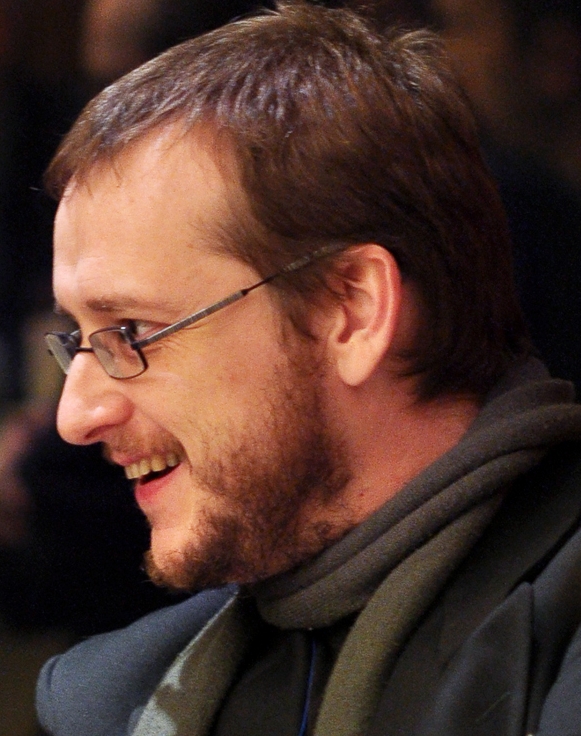}}]{Patrick Le Callet}
(F'19) received the M.Sc. and Ph.D. degrees in image processing from the Ecole Polytechnique de 1‘Universit\'{e} de Nantes. He was an Assistant Professor from 1997 to 1999 and a full time Lecturer from 1999 to 2003 with the Department of Electrical Engineering, Technical Institute of the University of Nantes. He led the Image and Video Communication Laboratory, CNRS IRCCyN, from 2006 to 2016, and was one of the five members of the Steering Board of CNRS, from 2013 to 2016. Since 2015, he has been the Scientific Director of the cluster Ouest Industries Cratives, a five-year program gathering over ten institutions (including three universities). Since 2017, he has been one of the seven members of the Steering Board of the CNRS LS2N Laboratory (450 researchers), as a Representative of Polytech Nantes. Since 2019, he has been the fellow of IEEE. He is mostly involved in research dealing with the application of human vision modeling in image and video processing. 

\end{IEEEbiography}







\end{document}

%% file: macros.tex
\usepackage{amsmath,amsfonts,bm,relsize,dsfont,color}
\usepackage{epsfig,graphicx,times,amssymb,booktabs,multirow} 
\usepackage{soul} 
\usepackage{url}

\usepackage{tabulary,array}
\newcolumntype{x}[1]{>{\centering\arraybackslash}p{#1pt}}

\usepackage{multirow}
\definecolor{MyDarkBlue}{rgb}{0,0.08,1}
\definecolor{MyDarkGreen}{rgb}{0.02,0.6,0.02}
\definecolor{MyDarkRed}{rgb}{0.8,0.02,0.02}
\definecolor{MyDarkOrange}{rgb}{0.40,0.2,0.02}
\definecolor{MyPurple}{RGB}{111,0,255}
\definecolor{MyRed}{rgb}{1.0,0.0,0.0}
\definecolor{MyGold}{rgb}{0.75,0.6,0.12}
\definecolor{MyDarkgray}{rgb}{0.66, 0.66, 0.66}
\definecolor{MyDarkCyan}{rgb}{0.05, 0.55, 0.45}
\definecolor{MyBlack}{rgb}{0., 0., 0.}
\definecolor{MyMagenta}{rgb}{1., 0., 1.}
\definecolor{BerkeleyYellow}{RGB}{255,204,41}
\definecolor{BerkeleyLightBlue}{RGB}{94,146,221}
\definecolor{BkDarkBlue}{rgb}{.05,.07,.353}

\newcommand{\gray}[1]{\textcolor{MyDarkgray}{#1}}

\def\ie{\textit{i.e.}}
\def\eg{\textit{e.g.}}
\def\etc{\textit{etc}}

\definecolor{NewGreen}{rgb}{.43,.67,.27}
\definecolor{NewOrange}{rgb}{.92,.49,.19}
\definecolor{NewBlue}{rgb}{.26,.44,.76}

%% file: figures.tex
\newcommand{\FigMotivationI}{
\begin{figure}
    \centering
    \vspace{-5pt}
    \includegraphics[width=0.85\linewidth]{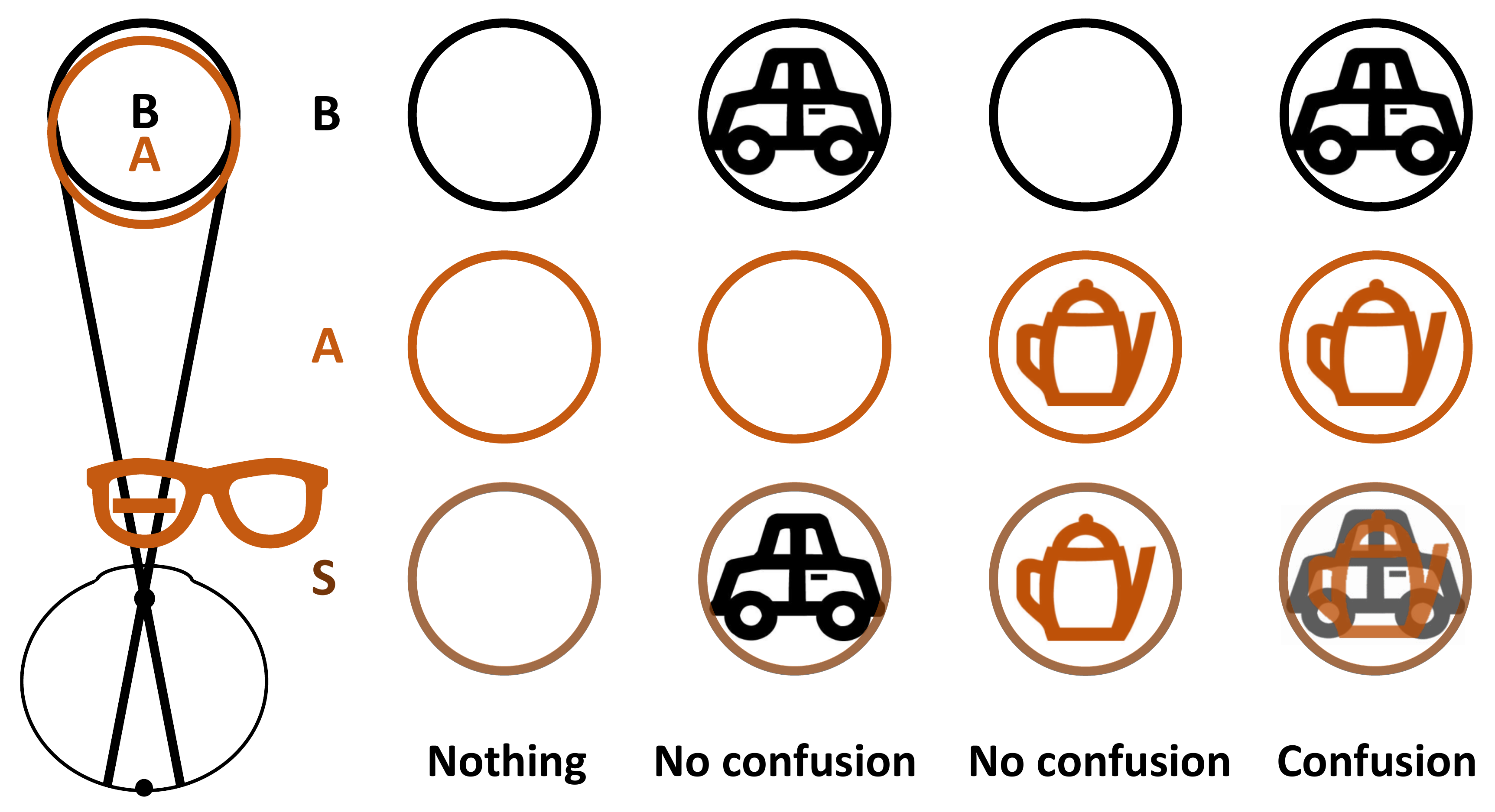}
    \vspace{-11pt}
    \caption{Visual confusion theory of AR. ``B'' denotes the background view, \textit{i.e.,} see-through view. ``A'' represents the augmented view. ``S'' implies the superimposition of the background view (B) and augmented view (A). ``A" and ``B" will influence the perceptual quality of each other.}
    \vspace{-13pt}
    \label{fig:1_visual_conf}
\end{figure}
}

\newcommand{\FigClarification}{
\begin{figure}
    \centering
    \includegraphics[width=0.98\linewidth]{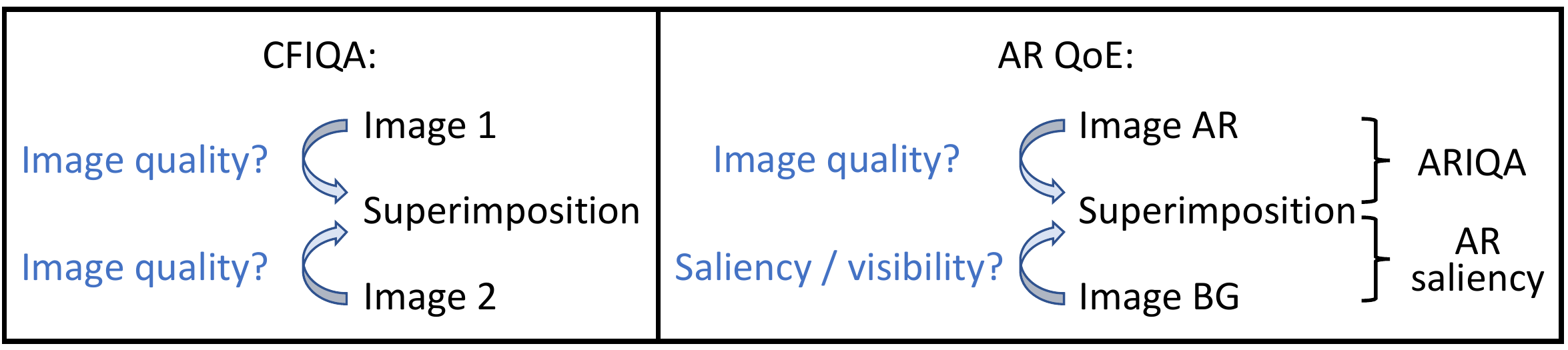}
    \vspace{-9pt}
    \caption{Relationship between CFIQA and ARIQA.}
    \label{fig:2_CFIQA_ARIQA}
    \vspace{-13pt}
\end{figure}
}

\newcommand{\FigDatasetI}{
\begin{figure}
    \centering
    \vspace{-2pt}
    \includegraphics[width=0.78\linewidth]{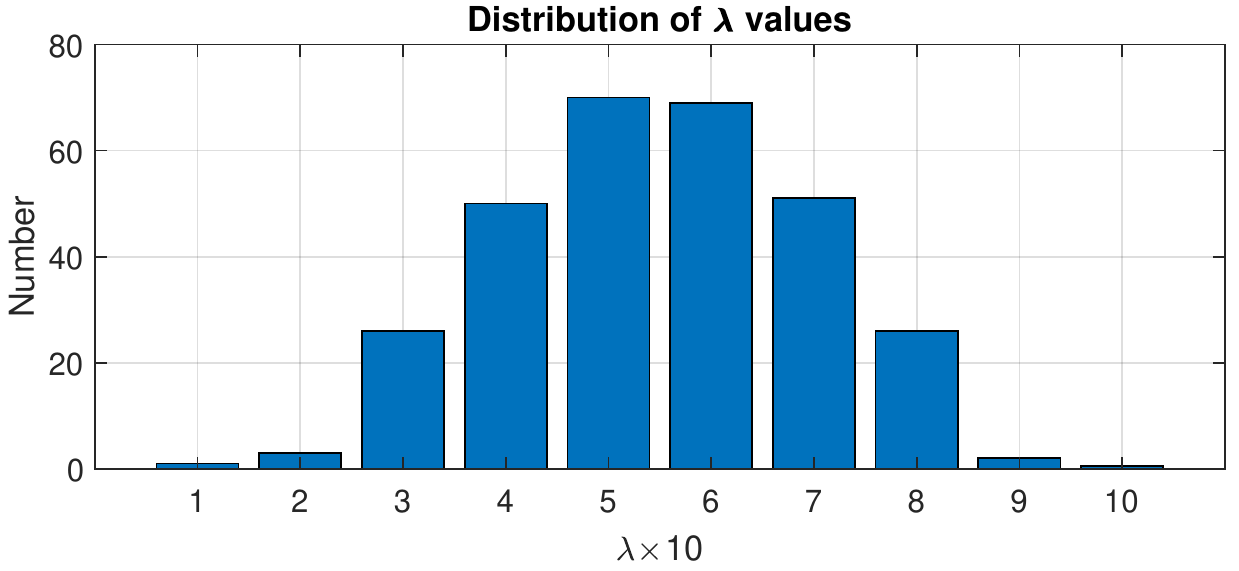}
    \vspace{-11pt}
    \caption{Distribution of $\lambda$ values in this paper. The values are multiplied by 10 and rounded up.}
    \label{fig:2_lambda}
    \vspace{-10pt}
\end{figure}
}

\newcommand{\FigDatasetII}{
\begin{figure}
    \centering
    \includegraphics[width=0.78\linewidth]{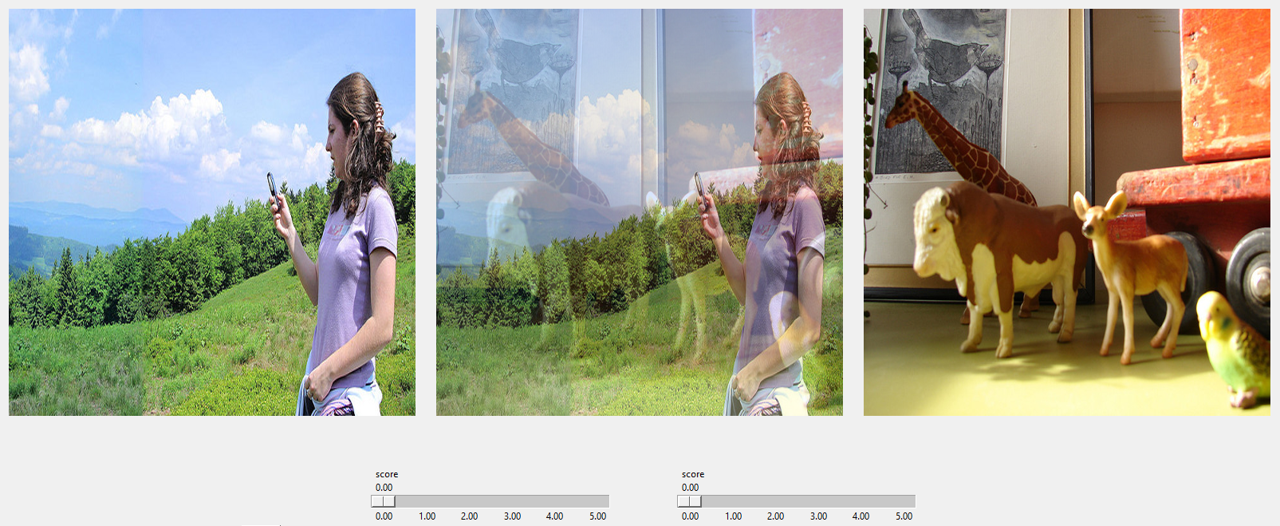}
    \vspace{-8pt}
    \caption{Illustration of the subjective experiment interface for CFIQA.}
    \vspace{-15pt}
    \label{fig:3_interface}
\end{figure}
}


\newcommand{\FigDatasetIII}{
\begin{figure}[t]
  \centering
  \includegraphics[width=0.8\linewidth]{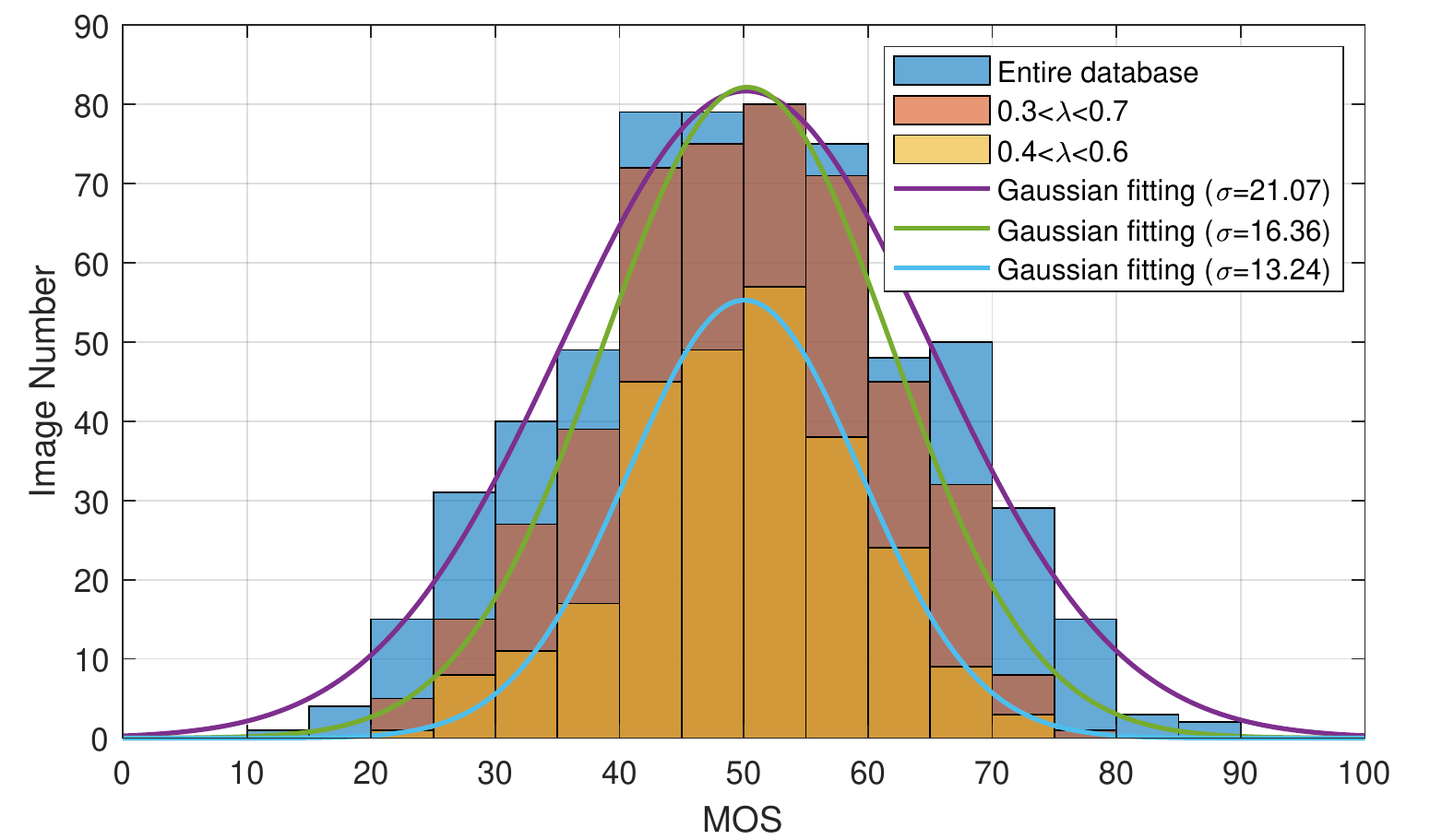}
  \vspace{-11pt}
  \caption{Histogram of MOSs from the CFIQA database within different $\lambda$ value ranges.}
  \vspace{-16pt}
  \label{fig:4_subjective}
\end{figure}
}

\newcommand{\FigDatasetIV}{
\begin{figure*}
    \centering
    \vspace{-3pt}
    \includegraphics[width=0.93\linewidth]{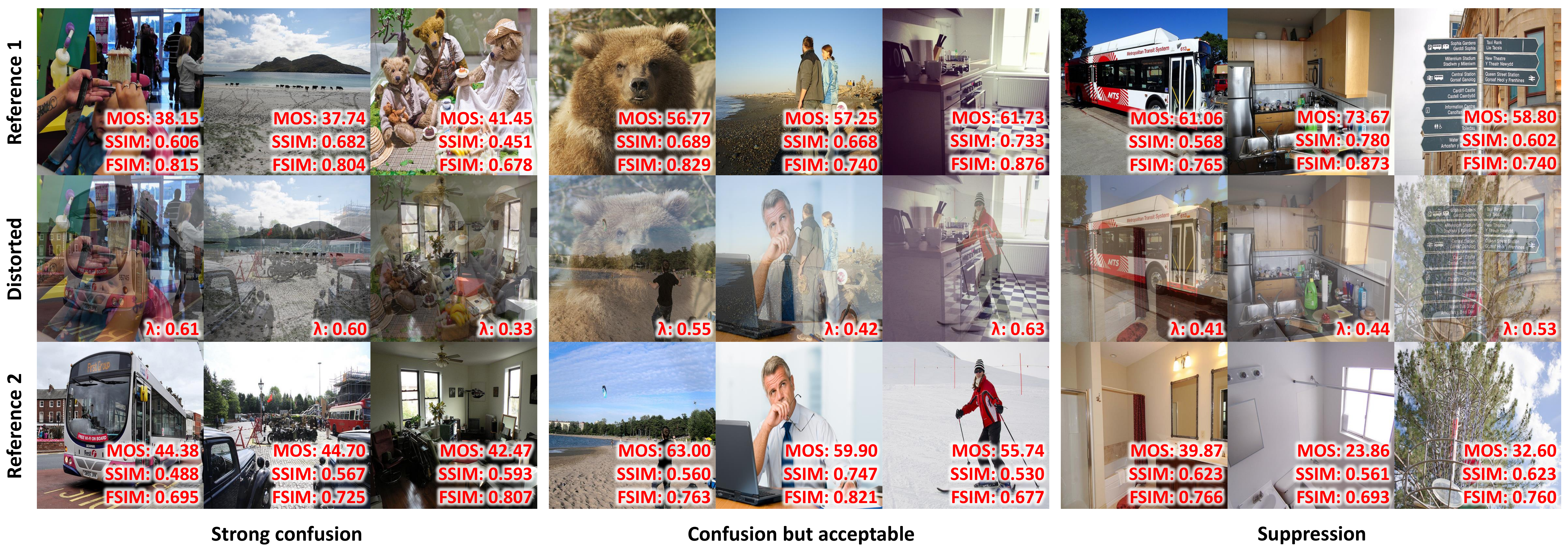}
    \vspace{-13pt}
    \caption{Sample images from CFIQA database. MOS, SSIM, FSIM values, as well as $\lambda$ value are given in the figure. Note that a MOS in this figure mean the MOS of the reference layer in the distorted image.}
    \vspace{-15pt}
    \label{fig:5_example}
\end{figure*}
}


\newcommand{\FigDatasetV}{
\begin{figure}[t]
  \centering
  \includegraphics[width=0.86\linewidth]{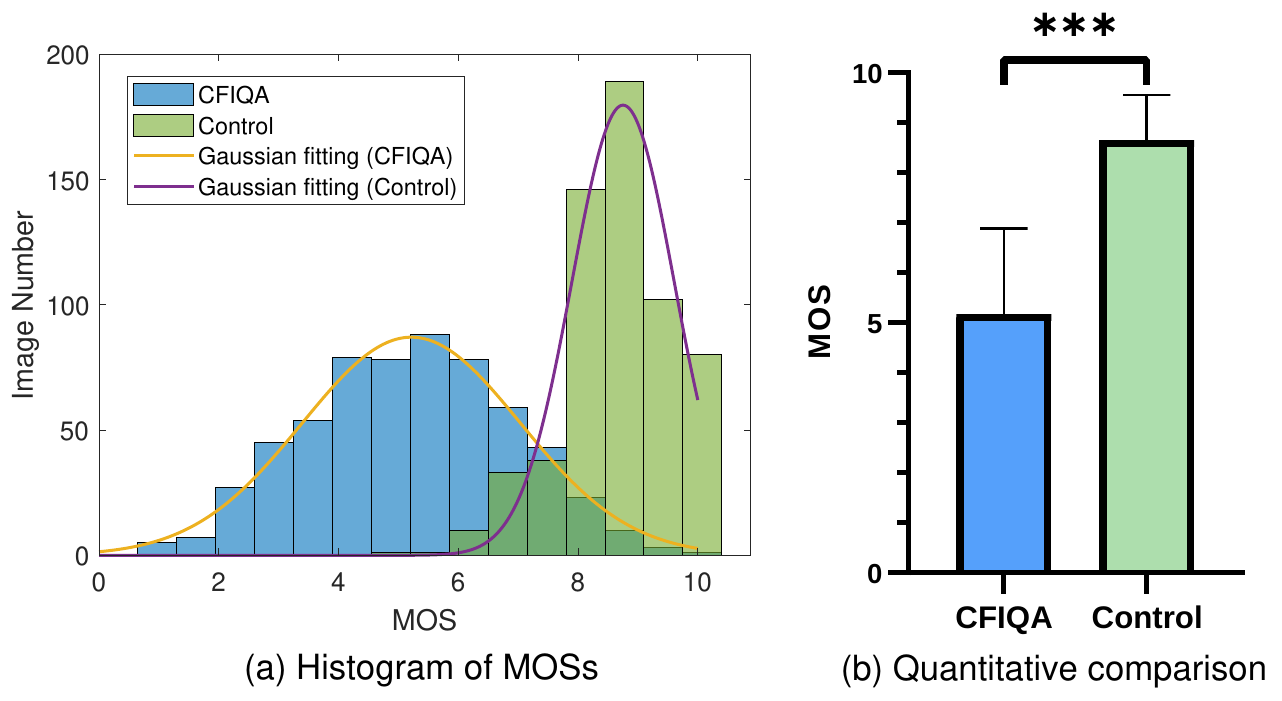}
  \vspace{-12pt}
  \caption{Comparisons of MOSs between the subjective CFIQA experiment and the controlled experiment.}
  \vspace{-16pt}
  \label{fig:5_control}
\end{figure}
}

\newcommand{\FigMethodI}{
\begin{figure*}
    \centering
    \includegraphics[width=0.9\linewidth]{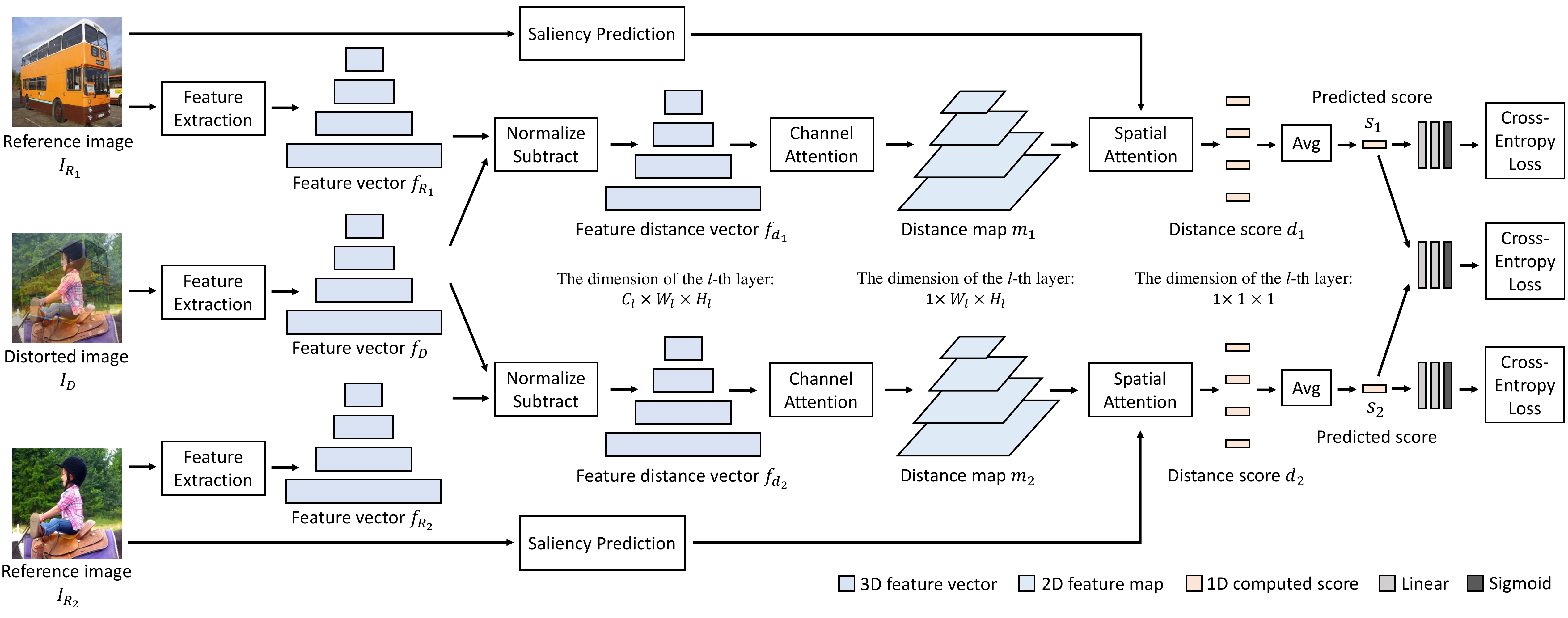}
    \vspace{-14pt}
    \caption{Illustration of our proposed attention based deep feature fusion method. 
    For one distorted image $I_D$ and two reference images $I_{R_1}$ and $I_{R_2}$, three DNN feature vectors $f_{D}$, $f_{R_1}$, and $f_{R_2}$ are first extracted.
    We then compute the feature distances between the corresponding feature layers of $f_{D}$ and $f_{R_1}$, as well as $f_{D}$ and $f_{R_2}$, respectively, to get distance vectors $f_{d_1}$, $f_{d_2}$.
    Next, two feature distance vectors $f_{d_1}$, $f_{d_2}$ are fed into a channel attention module to get distance map stacks $m_1$, $m_2$.
    After weighting by a spatial attention operation, two score vectors $d_1$ and $d_2$ are computed.
    Finally, two predicted scores $s_1$ and $s_2$ are calculated by averaging $d_1$ and $d_2$ respectively.
    Two kinds of loss functions are used to constrain the learning process, including the ranking loss (middle), and the score regression loss (top and bottom). All loss functions are based on the cross-entropy loss.
    }
    \vspace{-16pt}
    \label{fig:6_model}
\end{figure*}
}

\newcommand{\FigMethodCA}{
\begin{figure}
    \centering
    \includegraphics[width=0.7\linewidth]{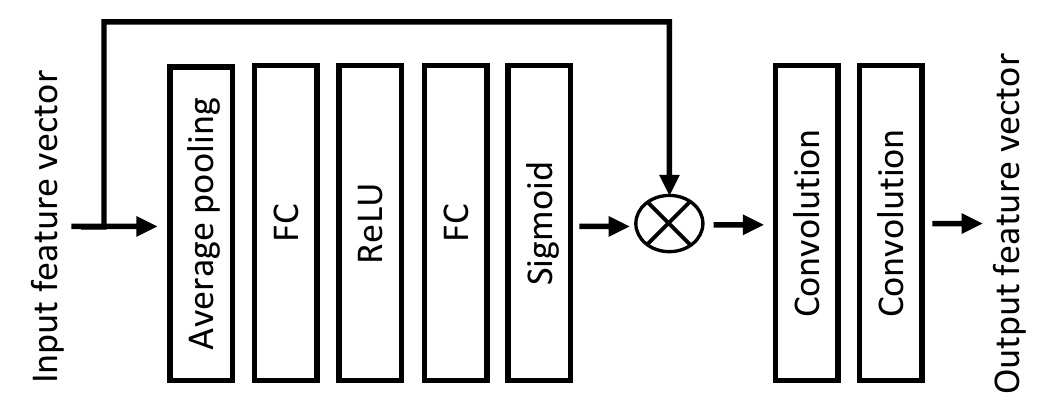}
    \vspace{-10pt}
    \caption{Illustration of the channel attention module used in the proposed CFIQA model.}
    \vspace{-10pt}
    \label{fig:7_modelCA}
\end{figure}
}

\newcommand{\FigVGG}{
\begin{figure}
    \centering
    \includegraphics[width=0.75\linewidth]{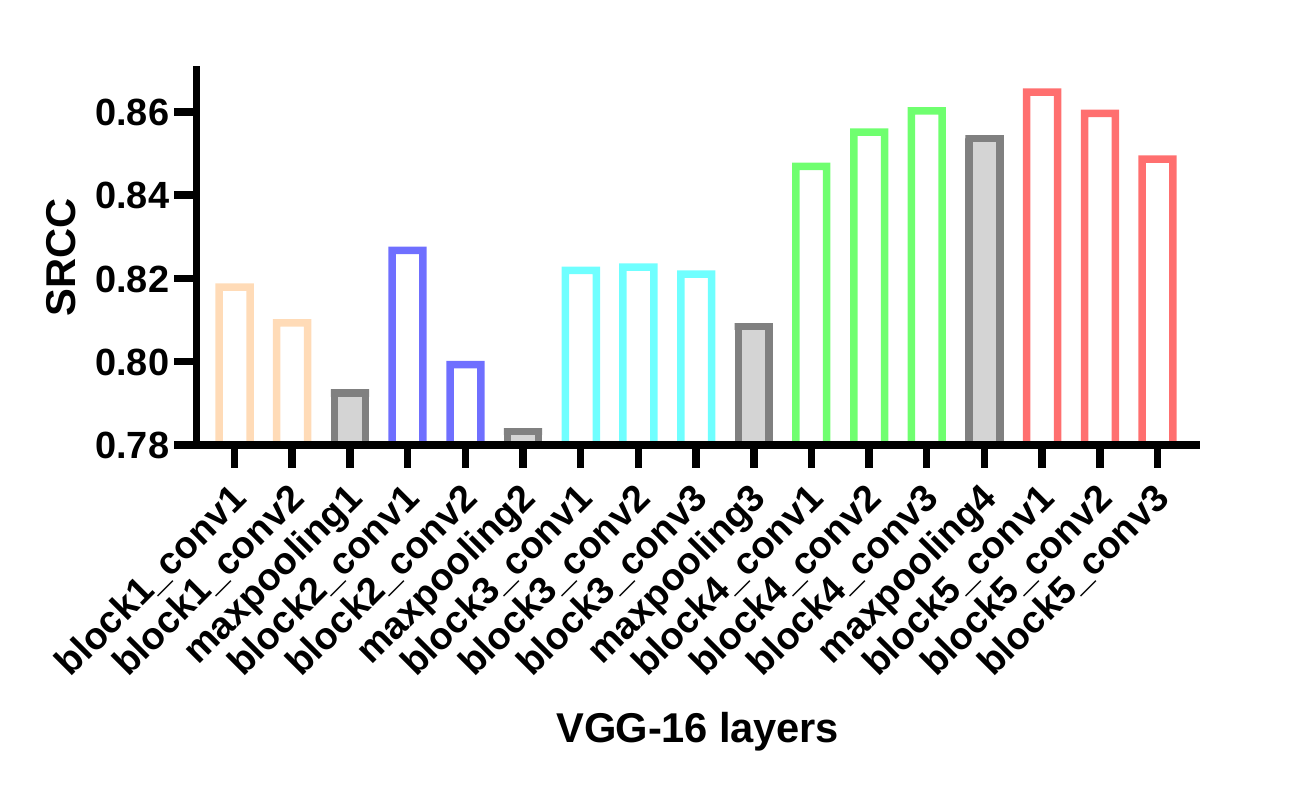}
    \vspace{-10pt}
    \caption{The SRCC scores between the distance scores and the subjective ratings for each layer $l$ of the VGG-16 network.}
    \vspace{-10pt}
    \label{fig:7_vgg_analysis}
\end{figure}
}

\newcommand{\FigARI}{
\begin{figure*}
\vspace{-10pt}
    \centering
    \includegraphics[width=0.75\linewidth]{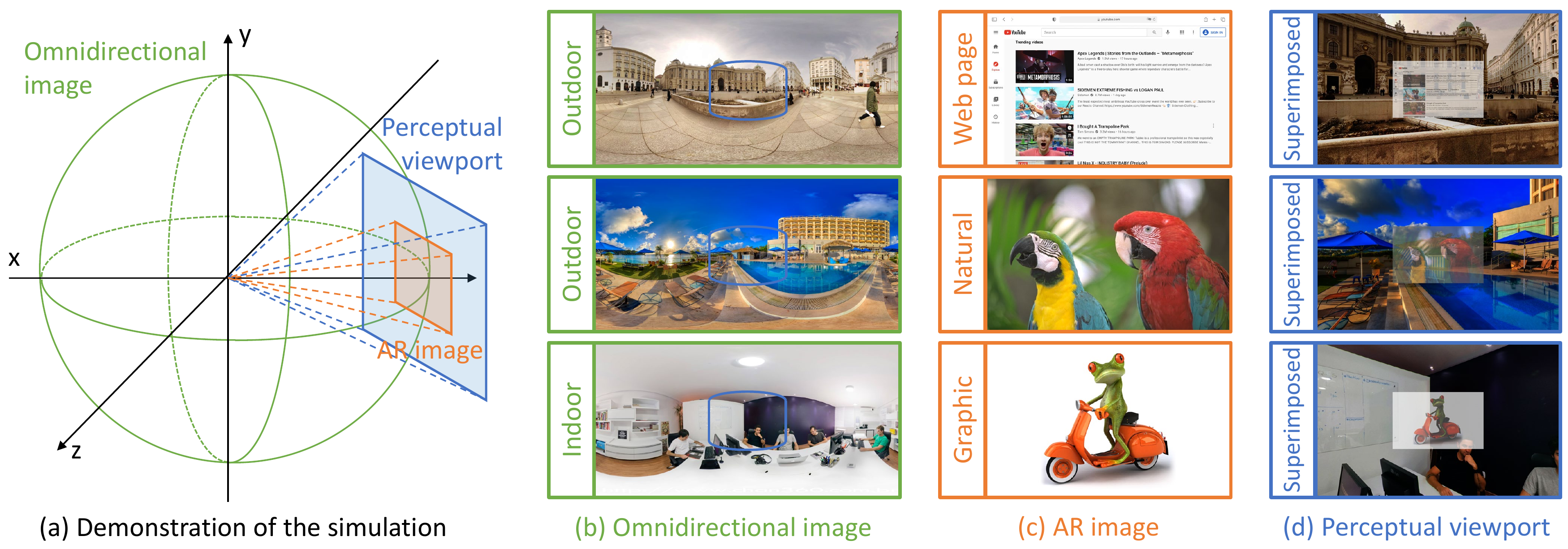}
    \vspace{-11pt}
    \caption{The illustration of the AR simulation in VR environment. (a) The demonstration of the relationship between the \textcolor{NewGreen}{omnidirectional image}, the \textcolor{NewOrange}{AR image}, and the \textcolor{NewBlue}{perceptual viewport image}. (b) The \textcolor{NewGreen}{omnidirectional images} are used as the background scenes, which include outdoor and indoor scenarios. (c) The \textcolor{NewOrange}{AR images} are composed of three types of content including web page images, natural images, and graphic images. (d) The \textcolor{NewBlue}{perceptual viewport images} are generated by superimposing the AR images on the omnidirectional images (here $\lambda=0.58$). Note that the perceptual viewports of the subjects are changed dynamically with the head movement, however, the relative positional relationship between the omnidirectional image and the AR image is fixed.}
    \vspace{-16pt}
    \label{fig:7_ar_simulation}
\end{figure*}
}

\newcommand{\FigARInterface}{
\begin{figure}
    \centering
    \includegraphics[width=0.85\linewidth]{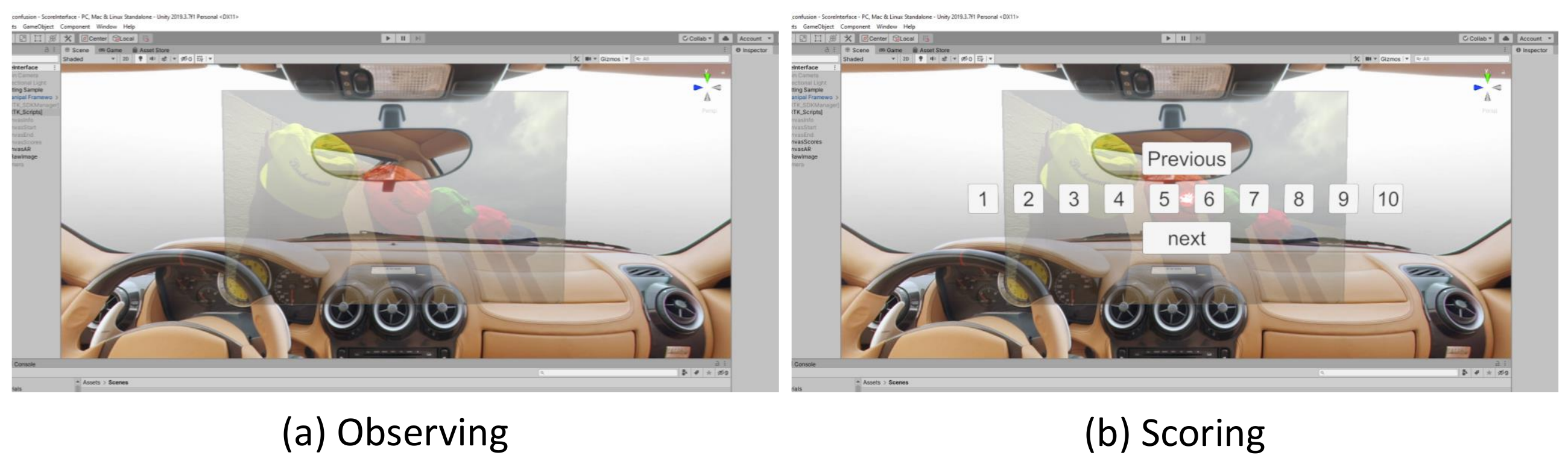}
    \vspace{-10pt}
    \caption{Demonstration of the subjective experiment interface for ARIQA. Note that this subjective experiment is conducted in VR environment, however this demonstration is the screenshot from the desktop.}
    \vspace{-12pt}
    \label{fig:7_ar_interface}
\end{figure}
}

\newcommand{\FigARII}{
\begin{figure}
    \centering
    \includegraphics[width=0.75\linewidth]{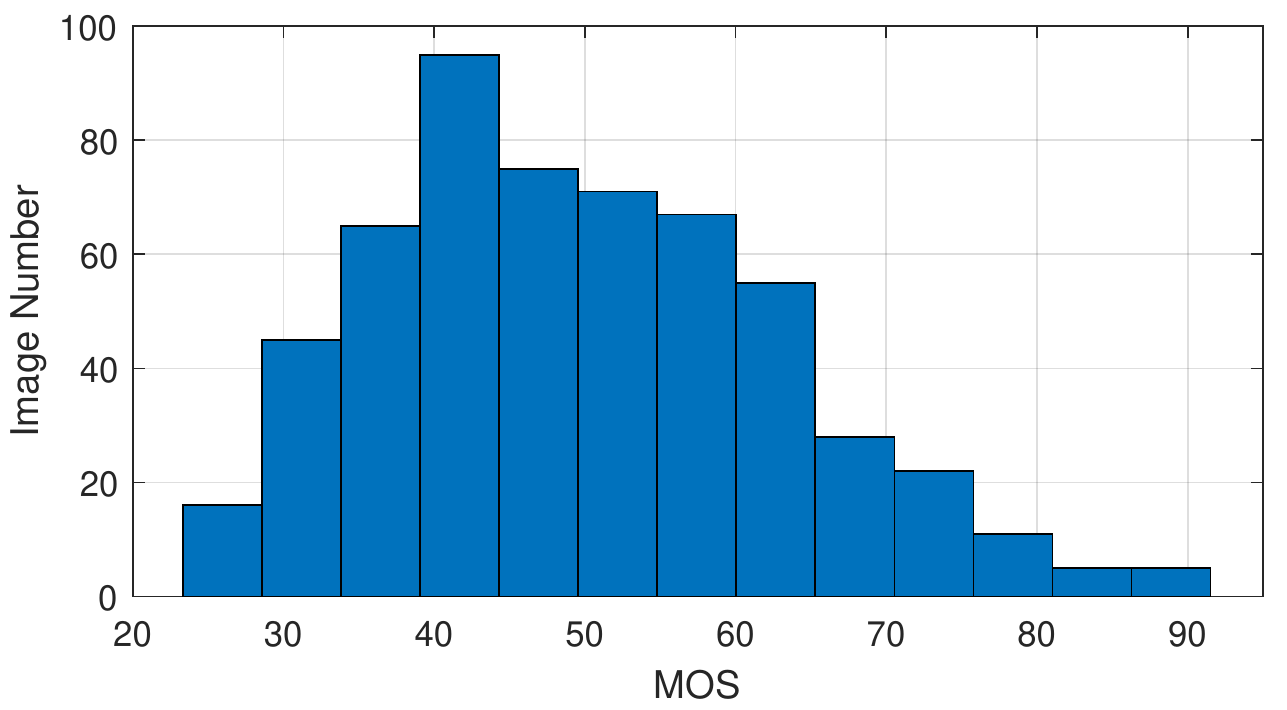}
    \vspace{-12pt}
    \caption{Histogram of MOSs from the ARIQA database.}
    \vspace{-16pt}
    \label{fig:8_ariqa_dataset}
\end{figure}
}

\newcommand{\FigARIII}{
\begin{figure}
\vspace{-2pt}
    \centering
    \includegraphics[width=0.88\linewidth]{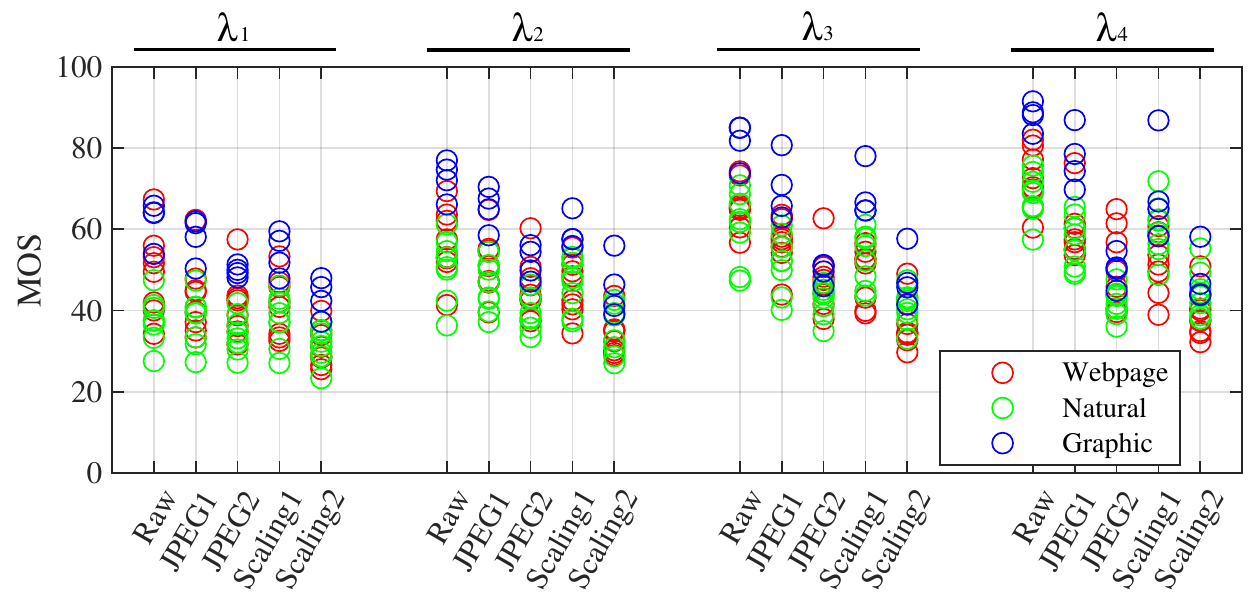}
    \vspace{-13pt}
    \caption{Distribution of MOS values of raw images, JPEG compressed images, rescaled images superimposed on the omnidirectional backgrounds with different mixing values. The mixing values $\lambda_1$, $\lambda_2$, $\lambda_3$, $\lambda_4$ are equal to 0.26, 0.42, 0.58, 0.74, respectively.}
    \vspace{-12pt}
    \label{fig:9_ariqa_dataset_analysis1}
\end{figure}
}

\newcommand{\FigARIV}{
\begin{figure}[t]
    \centering
    \includegraphics[width=0.88\linewidth]{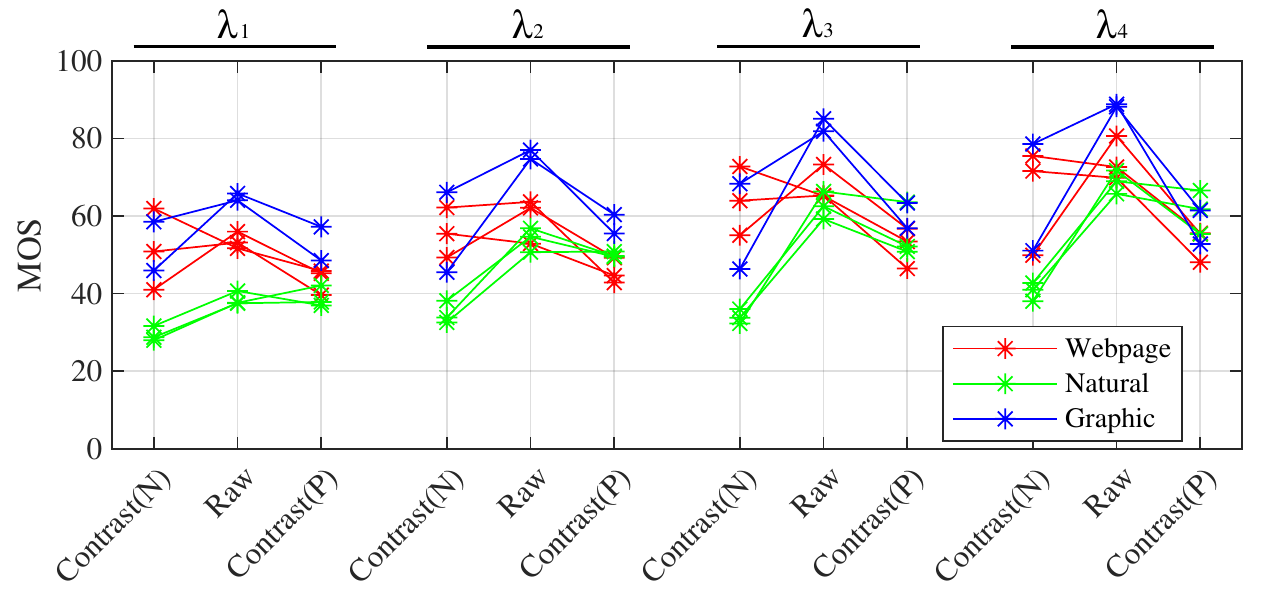}
    \vspace{-11pt}
    \caption{Samples of MOS values of raw images, contrast adjusted images superimposed on the omnidirectional backgrounds with different mixing values. ``N" denotes negative gamma transfer, ``P" represents positive gamma transfer.}
    \vspace{-16pt}
    \label{fig:10_ariqa_dataset_analysis2}
\end{figure}
}

\newcommand{\FigMethodII}{
\begin{figure}
\vspace{-2pt}
    \centering
    \includegraphics[width=0.85\linewidth]{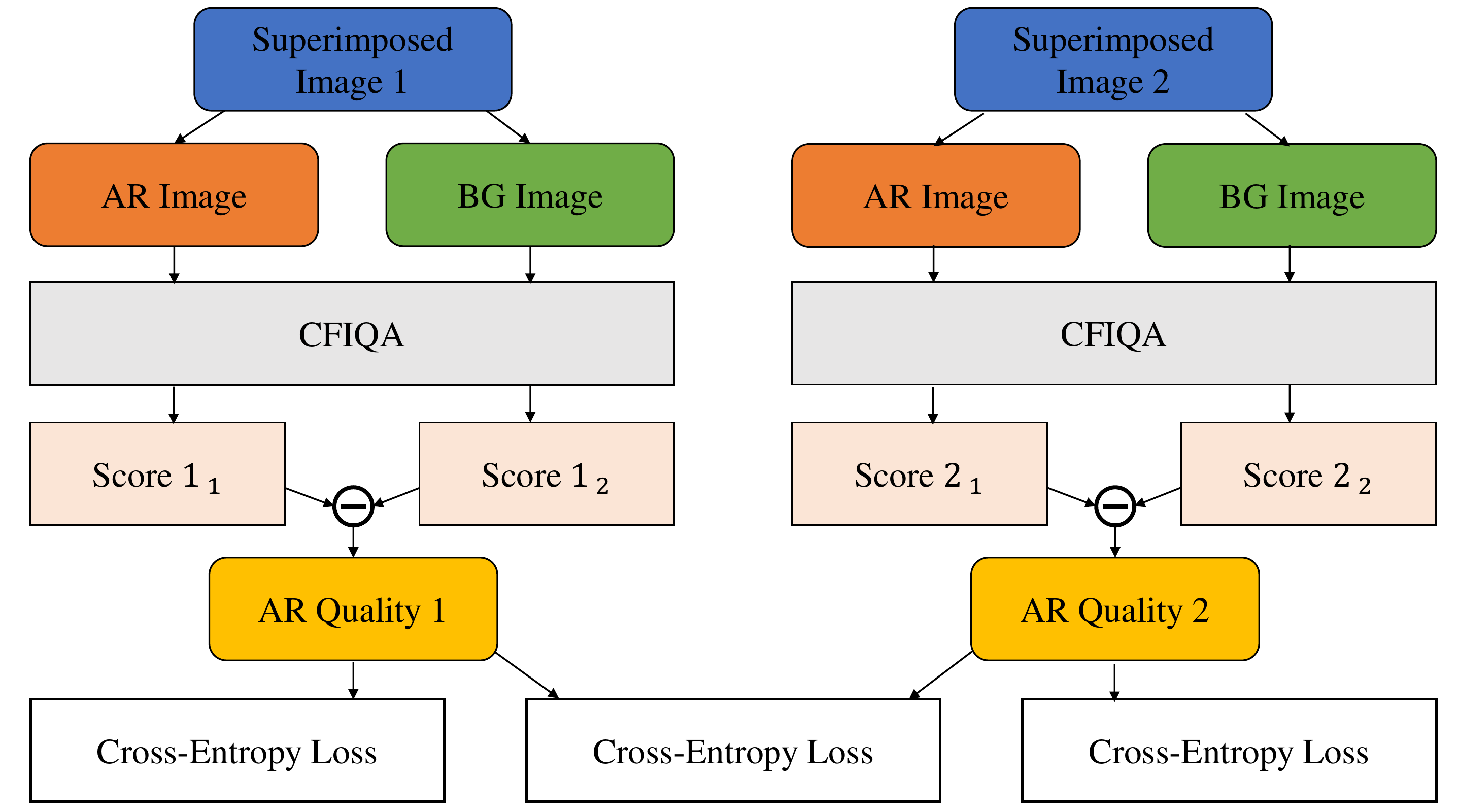}
    \vspace{-10pt}
    \caption{The framework of the proposed ARIQA model.}
    \vspace{-16pt}
    \label{fig:11_modelII}
\end{figure}
}

\newcommand{\FigNewCriterionI}{
\begin{figure*}[t]
\vspace{-4pt}
  \begin{minipage}{0.5\linewidth}
  \centering
  {\includegraphics[width=0.9\textwidth]{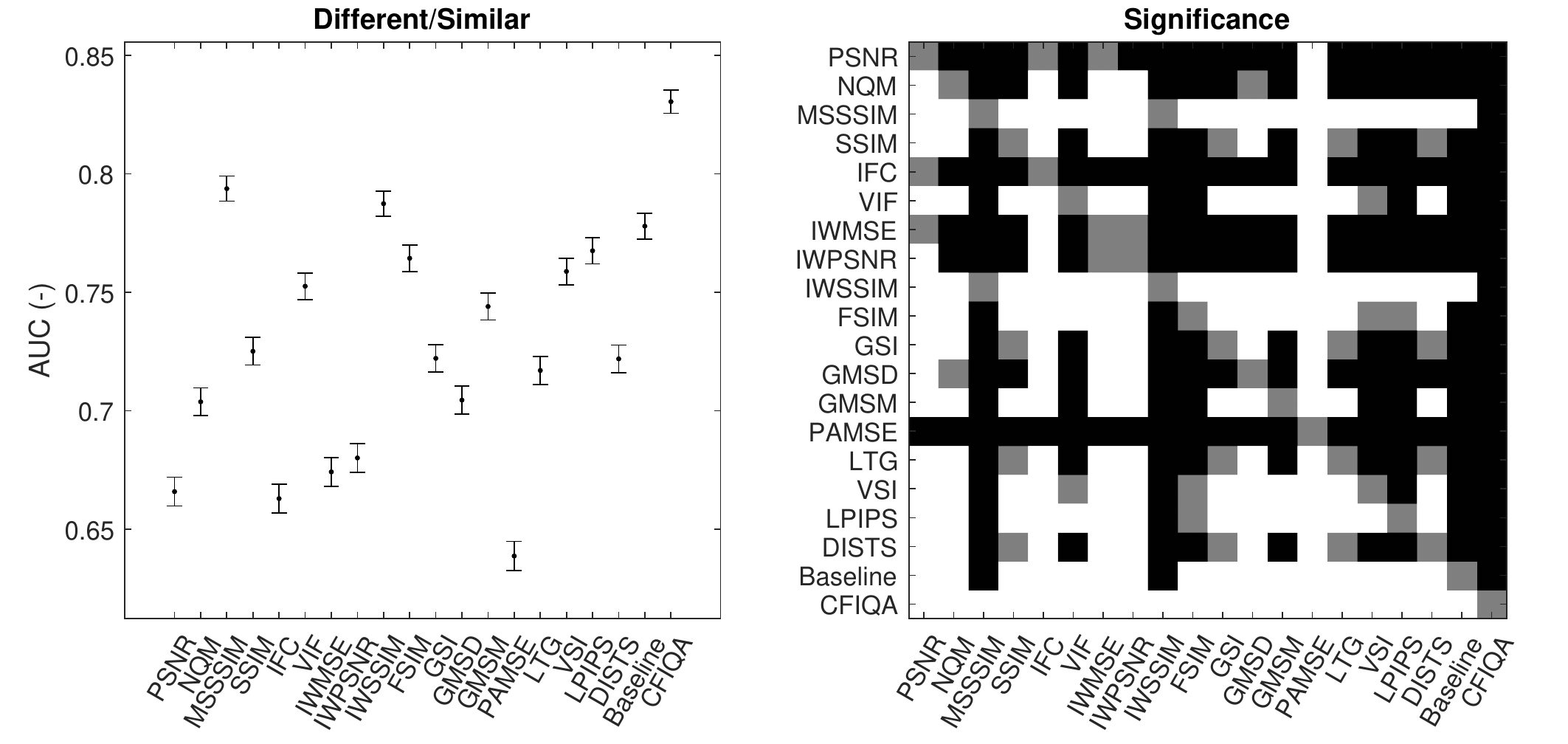}%
  \label{fig:new_evaluation_cfiqa_1}}
  \end{minipage}
  \hfil
  \begin{minipage}{0.5\linewidth}
  \centering
  {\includegraphics[width=0.9\textwidth]{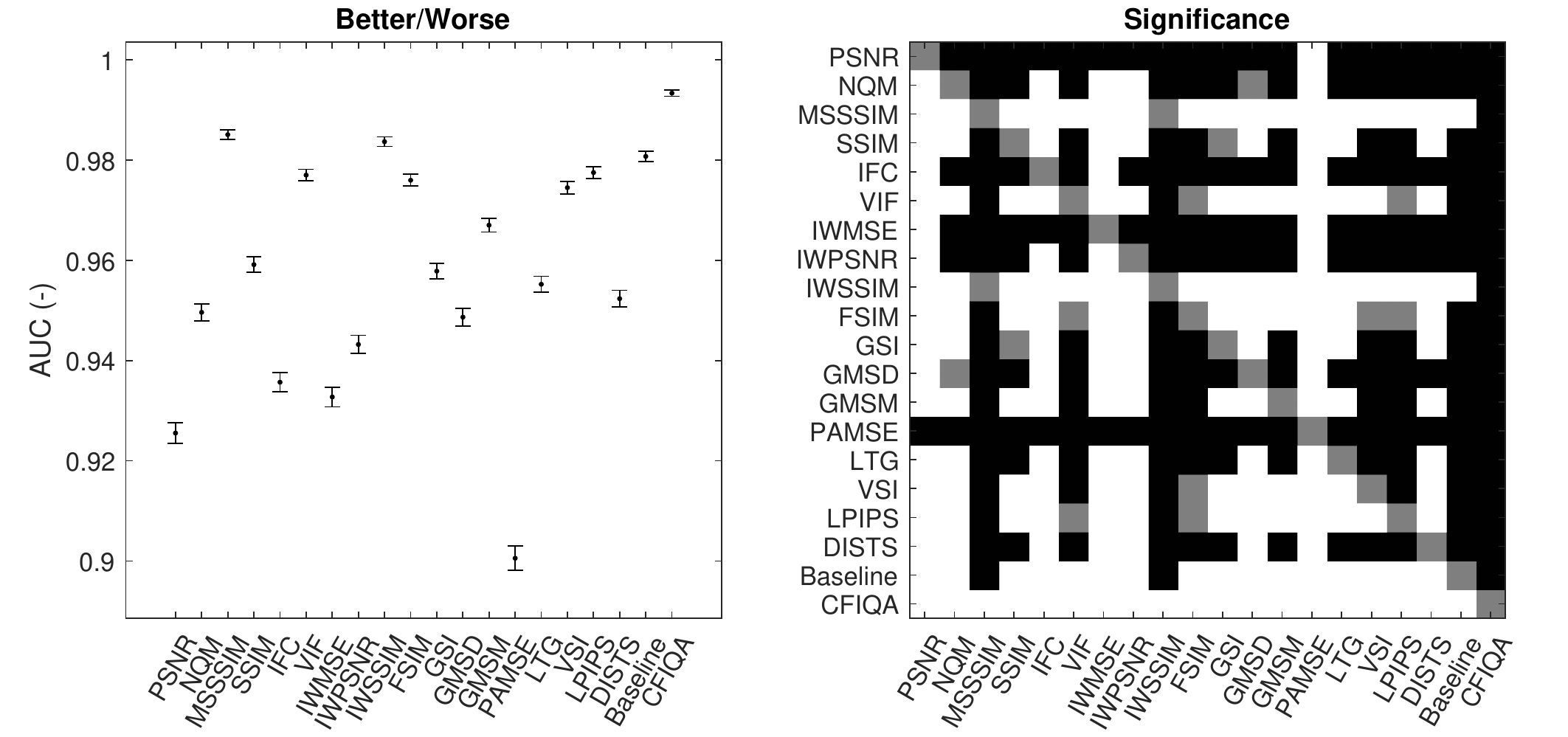}%
  \label{fig:new_evaluation_cfiqa_2}}
  \end{minipage}
  \vspace{-12pt}
  \caption{New criteria performance of 19 state-of-art FR IQA models and the proposed metric on the CFIQA database. Left two figures are the different vs. similar ROC analysis results. Right two figures are the better vs. worse analysis results. Note that a white/black square in the significance figures means the row metric is statistically better/worse than the column one. A gray square means the row method and the column method are statistically indistinguishable. The backbone of all networks in these figures is VGG-16.}
  \vspace{-14pt}
  \label{fig:new_evaluation_cfiqa}
\end{figure*}
}

\newcommand{\FigNewCriterionII}{
\begin{figure*}[t]
  \begin{minipage}{0.5\linewidth}
  \centering
  {\includegraphics[width=0.9\textwidth]{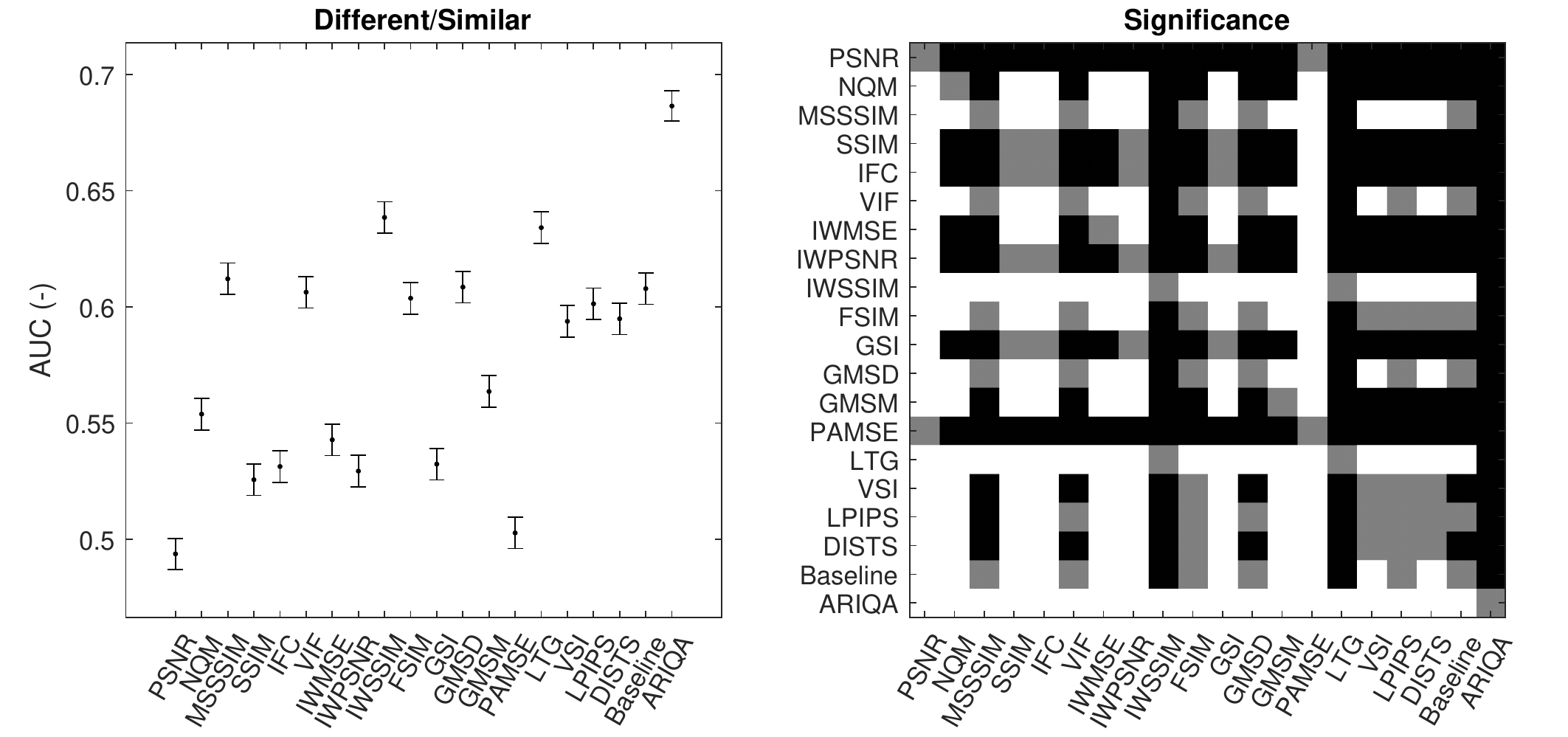}%
  \label{fig:new_evaluation_ariqa_1}}
  \end{minipage}
  \hfil
  \begin{minipage}{0.5\linewidth}
  \centering
  {\includegraphics[width=0.9\textwidth]{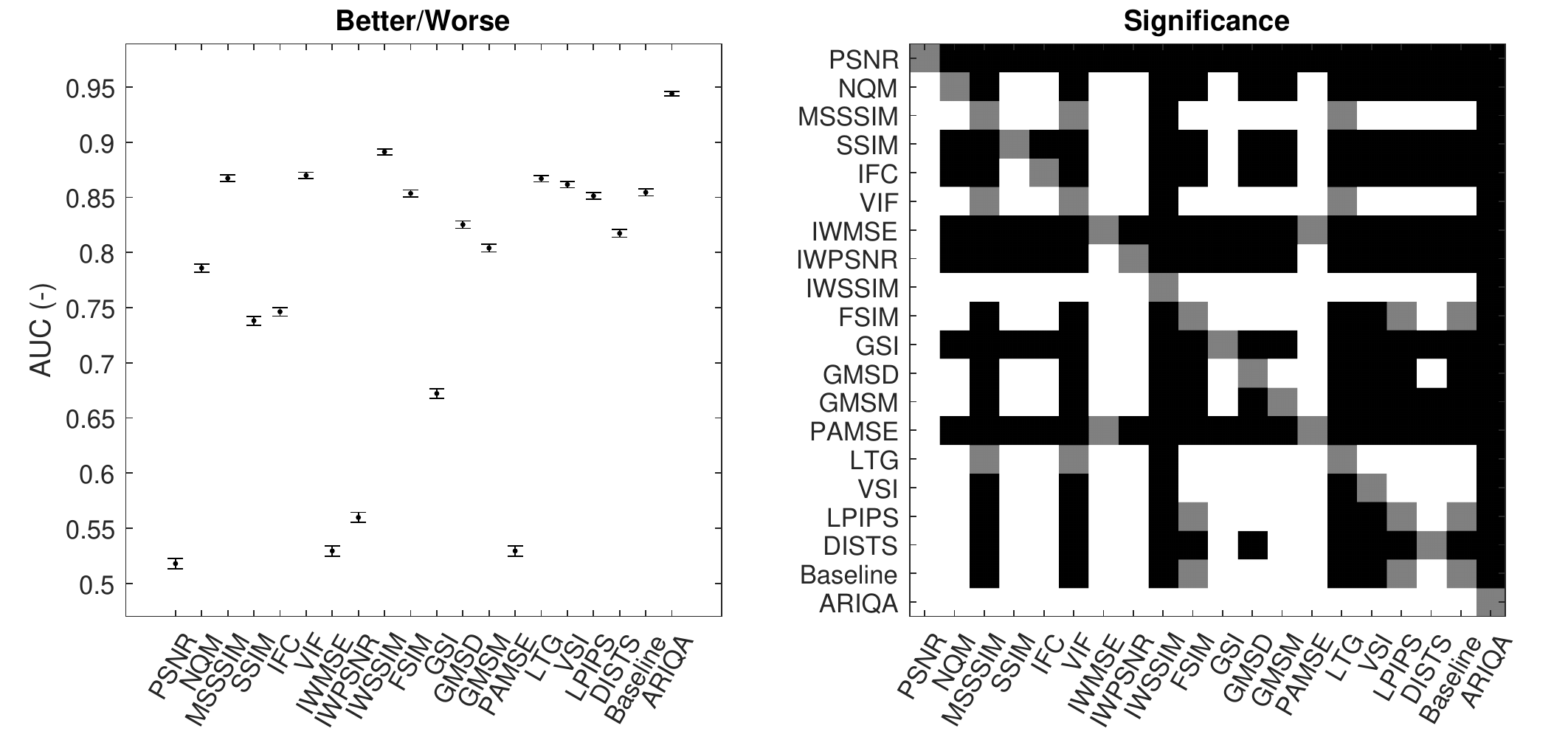}%
  \label{fig:new_evaluation_ariqa_2}}
  \end{minipage}
  \vspace{-10pt}
  \caption{New criteria performance of 19 state-of-art FR IQA models and the proposed metric on the ARIQA database. Left two figures are the different vs. similar ROC analysis results. Right two figures are the better vs. worse analysis results. The black/white/gray squares in the significance figures have the same meaning with that in Fig. \ref{fig:new_evaluation_cfiqa}}
  \vspace{-16pt}
  \label{fig:new_evaluation_ariqa}
\end{figure*}
}

%% file: tables.tex
\newcommand{\TabExperimentI}{
\begin{table*}
    \vspace{-10pt}
    \renewcommand{\arraystretch}{0.85}
    \setlength{\tabcolsep}{3pt}
    \fontsize{7.2pt}{\baselineskip}\selectfont
    \caption{Performance comparison of the state-of-the-art FR-IQA models on the CFIQA dataset. The dataset is split into three folds, including ``Entire'', ``0.3$<\lambda<$0.7'', and ``0.4$<\lambda<$0.6''. We \textbf{bold} the best two results in each group of models. ``*'' means a re-trained model.}
    \label{tab:1_all}
    \vspace{-5pt}
    \begin{tabular}{l@{\hskip 15pt} x{21.4}x{21.4}x{21.4}x{21.4}x{21.4} x{5} x{21.4}x{21.4}x{21.4}x{21.4}x{21.4} x{5} x{21.4}x{21.4}x{21.4}x{21.4}x{21.4}}
        \toprule
        Dataset & \multicolumn{5}{c}{Entire} && \multicolumn{5}{c}{0.3$<\lambda<$0.7} && \multicolumn{5}{c}{0.4$<\lambda<$0.6} \\
        \cline{2-6}\cline{8-12}\cline{14-18}
        \rule{0pt}{2.6ex}
        Model $\backslash$ Criteria & SRCC$\uparrow$ & KRCC$\uparrow$ & PLCC$\uparrow$ & RMSE$\downarrow$ & PWRC$\uparrow$ && SRCC$\uparrow$ & KRCC$\uparrow$ & PLCC$\uparrow$ & RMSE$\downarrow$ & PWRC$\uparrow$ && SRCC$\uparrow$ & KRCC$\uparrow$ & PLCC$\uparrow$ & RMSE$\downarrow$ & PWRC$\uparrow$\\
        \midrule 
        $\lambda$	&		0.7700	&	0.5881	&	0.8021	&	8.2909	&	11.130	&	&	0.5976	&	0.4283	&	0.6097	&	8.5426	&	7.6935	&	&	0.3597	&	0.2538	&	0.3698	&	8.6638	&	4.4961	\\
        \midrule																																
        MSE	&		0.6944	&	0.5072	&	0.7215	&	9.6118	&	9.8519	&	&	0.5030	&	0.3448	&	0.5073	&	9.3010	&	6.1129	&	&	0.2816	&	0.1909	&	0.3555	&	8.7250	&	3.3331	\\
        PSNR	&		0.6944	&	0.5072	&	0.7196	&	9.6399	&	9.8348	&	&	0.5030	&	0.3448	&	0.5076	&	9.2988	&	6.1341	&	&	0.2816	&	0.1909	&	0.3053	&	8.8574	&	3.3129	\\
        NQM \cite{damera2000image}	&		0.7486	&	0.5613	&	0.7683	&	8.8752	&	10.613	&	&	0.6012	&	0.4272	&	0.6131	&	8.5048	&	7.4643	&	&	0.4742	&	0.3282	&	0.5150	&	7.9967	&	5.4664	\\
        SSIM \cite{wang2004image}	&		0.7907	&	0.6005	&	0.8179	&	7.9970	&	11.204	&	&	0.6520	&	0.4660	&	0.6707	&	8.0269	&	8.0014	&	&	0.4858	&	0.3394	&	0.5143	&	7.9994	&	5.1510	\\
        IFC \cite{sheikh2005information}	&		0.7198	&	0.5362	&	0.7439	&	9.2430	&	10.461	&	&	0.5971	&	0.4253	&	0.6118	&	8.5491	&	7.4499	&	&	0.5416	&	0.3881	&	0.5610	&	7.6933	&	5.8762	\\
        VIF \cite{sheikh2006image}	&		\textbf{0.8346}	&	\textbf{0.6491}	&	\textbf{0.8556}	&	\textbf{7.1762}	&	\textbf{11.811}	&	&	\textbf{0.7266}	&	\textbf{0.5349}	&	\textbf{0.7372}	&	\textbf{7.3064}	&	\textbf{8.8275}	&	&	0.6244	&	\textbf{0.4548}	&	\textbf{0.6505}	&	\textbf{7.0875}	&	6.5839	\\
        IW-MSE \cite{wang2010information}	&		0.7468	&	0.5563	&	0.7777	&	8.7497	&	10.592	&	&	0.5787	&	0.4019	&	0.5999	&	8.6496	&	7.0436	&	&	0.3524	&	0.2348	&	0.3865	&	8.6132	&	3.6707	\\
        IW-PSNR \cite{wang2010information}	&		0.7468	&	0.5563	&	0.7694	&	8.8837	&	10.571	&	&	0.5787	&	0.4019	&	0.6008	&	8.6407	&	7.0474	&	&	0.3524	&	0.2348	&	0.3871	&	8.6105	&	3.5713	\\
        IW-SSIM \cite{wang2010information}	&		\textbf{0.8519}	&	\textbf{0.6715}	&	\textbf{0.8700}	&	\textbf{6.8096}	&	\textbf{12.015}	&	&	\textbf{0.7512}	&	\textbf{0.5607}	&	\textbf{0.7638}	&	\textbf{6.9567}	&	\textbf{9.0641}	&	&	\textbf{0.6633}	&	\textbf{0.4804}	&	\textbf{0.6856}	&	\textbf{6.7796}	&	\textbf{7.0594}	\\
        FSIM \cite{zhang2011fsim}	&		0.8331	&	0.6486	&	0.8540	&	7.2151	&	11.721	&	&	0.7201	&	0.5268	&	0.7263	&	7.4205	&	8.6681	&	&	\textbf{0.6296}	&	0.4501	&	0.6496	&	7.0917	&	\textbf{6.7445}	\\
        GSI \cite{liu2011image}	&		0.7904	&	0.6002	&	0.8152	&	8.0506	&	11.147	&	&	0.6507	&	0.4623	&	0.6656	&	8.0538	&	7.8943	&	&	0.5085	&	0.3516	&	0.5447	&	7.8056	&	5.5417	\\
        GMSD \cite{xue2013gradient}	&		0.7609	&	0.5725	&	0.7894	&	8.5356	&	10.812	&	&	0.6343	&	0.4533	&	0.6472	&	8.2345	&	7.7938	&	&	0.5678	&	0.3986	&	0.6152	&	7.3588	&	6.2830	\\
        GMSM \cite{xue2013gradient}	&		0.8117	&	0.6238	&	0.8328	&	7.6853	&	11.433	&	&	0.6930	&	0.5013	&	0.7022	&	7.6849	&	8.3621	&	&	0.5866	&	0.4122	&	0.6070	&	7.4067	&	6.3127	\\
        PAMSE \cite{xue2013perceptual}	&		0.6827	&	0.4965	&	0.7104	&	9.7694	&	9.6744	&	&	0.4855	&	0.3323	&	0.4930	&	9.3900	&	5.8931	&	&	0.2648	&	0.1802	&	0.3306	&	8.8155	&	3.0792	\\
        LTG \cite{gu2014efficient}	&		0.7922	&	0.6049	&	0.8160	&	8.0271	&	11.189	&	&	0.6672	&	0.4806	&	0.6779	&	7.9397	&	8.1256	&	&	0.5731	&	0.4071	&	0.6121	&	7.3762	&	6.3452	\\
        VSI \cite{zhang2014vsi}	&		0.8297	&	0.6455	&	0.8544	&	7.2150	&	11.679	&	&	0.7106	&	0.5188	&	0.7256	&	7.4266	&	8.6095	&	&	0.5860	&	0.4175	&	0.6285	&	7.2427	&	6.3984	\\
        \midrule																																		
        SSIM \cite{wang2004image} + saliency	&		\textbf{0.8364}	&	\textbf{0.6520}	&	\textbf{0.8528}	&	\textbf{7.2437}	&	\textbf{11.739}	&	&	\textbf{0.7228}	&	\textbf{0.5318}	&	\textbf{0.7362}	&	\textbf{7.3157}	&	\textbf{8.6880}	&	&	0.5797	&	0.4115	&	0.6063	&	7.4211	&	6.0386	\\
        FSIM \cite{zhang2011fsim} + saliency	&		\textbf{0.8542}	&	\textbf{0.6705}	&	\textbf{0.8711}	&	\textbf{6.8162}	&	\textbf{11.926}	&	&	\textbf{0.7558}	&	\textbf{0.5584}	&	\textbf{0.7664}	&	\textbf{6.9428}	&	\textbf{8.9966}	&	&	\textbf{0.6628}	&	\textbf{0.4724}	&	\textbf{0.6728}	&	\textbf{6.9110}	&	\textbf{6.9979}	\\
        GMSM \cite{xue2013gradient} + saliency	&		0.8339	&	0.6470	&	0.8519	&	7.2608	&	11.672	&	&	0.7227	&	0.5268	&	0.7336	&	7.3292	&	8.5930	&	&	\textbf{0.6176}	&	\textbf{0.4354}	&	\textbf{0.6223}	&	\textbf{7.2978}	&	\textbf{6.4243}	\\
        \midrule																																		
        LPIPS (Squeeze) \cite{zhang2018unreasonable}	&		0.8338	&	0.6468	&	0.8521	&	7.2596	&	11.676	&	&	0.7278	&	0.5341	&	0.7418	&	7.2346	&	8.7265	&	&	0.6117	&	0.4399	&	0.6708	&	6.9190	&	6.5814	\\
        LPIPS (Alex) \cite{zhang2018unreasonable}	&		0.8474	&	0.6610	&	0.8612	&	7.0603	&	11.847	&	&	0.7546	&	0.5615	&	0.7647	&	6.9547	&	9.0461	&	&	0.6583	&	0.4793	&	0.6943	&	6.7094	&	7.2792	\\
        LPIPS (VGG) \cite{zhang2018unreasonable}	&		0.8376	&	0.6512	&	0.8520	&	7.2607	&	11.762	&	&	0.7439	&	0.5501	&	0.7535	&	7.0982	&	8.9189	&	&	0.6551	&	0.4772	&	0.7013	&	6.6587	&	6.9437	\\
        LPIPS (VGG) \cite{zhang2018unreasonable} *	&		0.8413	&	0.6549	&	0.8607	&	7.0817	&	11.799	&	&	0.7508	&	0.5550	&	0.7596	&	7.0379	&	8.9663	&	&	0.6673	&	0.4881	&	0.6950	&	6.7167	&	7.0033	\\
        DISTS \cite{ding2020image}	&		0.7709	&	0.5815	&	0.7967	&	8.4070	&	10.684	&	&	0.6526	&	0.4658	&	0.6810	&	7.9190	&	7.7870	&	&	0.5253	&	0.3665	&	0.5531	&	7.7848	&	5.7042	\\
        Baseline (SqueezeNet)	&		0.8625	&	0.6806	&	0.8774	&	6.6574	&	12.038	&	&	0.7760	&	0.5823	&	0.7822	&	6.7219	&	9.2077	&	&	0.6902	&	0.5046	&	0.7170	&	6.5044	&	7.4181	\\
        Baseline (AlexNet)	&		0.8752	&	0.6987	&	0.8870	&	6.4044	&	12.189	&	&	0.7967	&	0.6061	&	0.8049	&	6.4086	&	9.4289	&	&	0.7162	&	0.5321	&	0.7364	&	6.3125	&	7.4786	\\
        Baseline (VGG-16)	&		0.8494	&	0.6655	&	0.8663	&	6.9274	&	11.944	&	&	0.7630	&	0.5683	&	0.7712	&	6.8757	&	9.1520	&	&	0.6792	&	0.4970	&	0.7108	&	6.5686	&	7.4597	\\
        Baseline+ (VGG-16)	&		0.8629	&	0.6820	&	0.8772	&	6.6615	&	12.078	&	&	0.7847	&	0.5910	&	0.7921	&	6.5973	&	9.3582	&	&	0.7028	&	0.5167	&	0.7265	&	6.4195	&	7.4188	\\
        Baseline (VGG-19)	&		0.8622	&	0.6798	&	0.8769	&	6.6573	&	12.106	&	&	0.7804	&	0.5847	&	0.7859	&	6.6754	&	9.3364	&	&	0.7010	&	0.5126	&	0.7327	&	6.3490	&	7.4755	\\
        Baseline (ResNet-18)	&		\textbf{0.8784}	&	\textbf{0.7004}	&	\textbf{0.8920}	&	\textbf{6.2736}	&	\textbf{12.236}	&	&	\textbf{0.8025}	&	\textbf{0.6079}	&	\textbf{0.8087}	&	\textbf{6.3598}	&	\textbf{9.5158}	&	&	0.7227	&	0.5318	&	0.7393	&	6.2907	&	7.4336	\\
        Baseline (ResNet-34)	&		\textbf{0.8823}	&	\textbf{0.7053}	&	\textbf{0.8941}	&	\textbf{6.2185}	&	\textbf{12.273}	&	&	\textbf{0.8098}	&	\textbf{0.6157}	&	\textbf{0.8137}	&	\textbf{6.2847}	&	\textbf{9.5477}	&	&	\textbf{0.7405}	&	\textbf{0.5490}	&	\textbf{0.7570}	&	\textbf{6.1058}	&	\textbf{7.6783}	\\
        Baseline (ResNet-50)	&		0.8759	&	0.6963	&	0.8885	&	6.3699	&	12.215	&	&	0.8011	&	0.6060	&	0.8060	&	6.3992	&	9.4857	&	&	\textbf{0.7250}	&	\textbf{0.5340}	&	\textbf{0.7447}	&	\textbf{6.2347}	&	\textbf{7.5612}	\\
            \midrule																																		
        CFIQA (SqueezeNet)	&		0.8712	&	0.6933	&	0.8827	&	6.5180	&	12.112	&	&	0.7847	&	0.5910	&	0.7933	&	6.5641	&	9.2214	&	&	0.6904	&	0.5050	&	0.6994	&	6.6646	&	6.9383	\\
        CFIQA (AlexNet)	&		0.9053	&	0.7347	&	0.9109	&	5.7467	&	12.466	&	&	0.8438	&	0.6520	&	0.8461	&	5.7758	&	9.7944	&	&	0.7749	&	0.5833	&	0.7878	&	5.7559	&	7.8676	\\
        CFIQA (VGG-16)	&		0.9203	&	0.7550	&	\textbf{0.9258}	&	\textbf{5.2636}	&	\textbf{12.652}	&	&	0.8681	&	0.6782	&	0.8700	&	5.3412	&	10.012	&	&	0.8055	&	0.6105	&	0.8168	&	5.3859	&	8.1268	\\
        CFIQA (VGG-19)	&		0.9178	&	0.7527	&	0.9229	&	5.3597	&	12.611	&	&	0.8643	&	0.6762	&	0.8646	&	5.4443	&	9.9785	&	&	0.7987	&	0.6068	&	0.8119	&	5.4520	&	8.0631	\\
        CFIQA (ResNet-18)	&		0.9196	&	0.7580	&	0.9256	&	5.2713	&	12.632	&	&	0.8677	&	0.6844	&	0.8688	&	5.3661	&	9.9947	&	&	0.8118	&	0.6272	&	0.8291	&	5.2225	&	8.1516	\\
        CFIQA (ResNet-34)	&		\textbf{0.9209}	&	\textbf{0.7604}	&	0.9256	&	5.2746	&	12.645	&	&	\textbf{0.8703}	&	\textbf{0.6883}	&	\textbf{0.8706}	&	\textbf{5.3293}	&	\textbf{10.032}	&	&	\textbf{0.8204}	&	\textbf{0.6381}	&	\textbf{0.8310}	&	\textbf{5.1876}	&	\textbf{8.2088}	\\
        CFIQA+ (ResNet-34)	&		\textbf{0.9232}	&	\textbf{0.7625}	&	\textbf{0.9278}	&	\textbf{5.1966}	&	\textbf{12.668}	&	&	\textbf{0.8733}	&	\textbf{0.6897}	&	\textbf{0.8741}	&	\textbf{5.2624}	&	\textbf{10.058}	&	&	\textbf{0.8226}	&	\textbf{0.6341}	&	\textbf{0.8326}	&	\textbf{5.1592}	&	\textbf{8.2436}	\\
        CFIQA (ResNet-50)	&		0.8964	&	0.7241	&	0.9042	&	5.9441	&	12.402	&	&	0.8275	&	0.6370	&	0.8311	&	6.0228	&	9.6863	&	&	0.7664	&	0.5774	&	0.7929	&	5.6932	&	7.9248	\\
        \bottomrule
    \end{tabular}
    \vspace{-15pt}
\end{table*}
}

\newcommand{\TabExperimentII}{
\begin{table*}
    \renewcommand{\arraystretch}{0.84}
    \setlength{\tabcolsep}{3pt}
    \fontsize{7.2pt}{\baselineskip}\selectfont
    \caption{Impact of different components. (Backbone: VGG-16. ``CA'' means ``channel attention''. ``SA'' means ``spatial attention'').}
    \vspace{-7pt}
    \label{tab:2_ablation}
  
    \begin{tabular}{l@{\hskip 18pt} x{22}x{22}x{22}x{22}x{22} x{6} x{22}x{22}x{22}x{22}x{22} x{6} x{22}x{22}x{22}x{22}x{22}}
    
        \toprule
        Dataset & \multicolumn{5}{c}{Entire} && \multicolumn{5}{c}{0.3$<\lambda<$0.7} && \multicolumn{5}{c}{0.4$<\lambda<$0.6} \\
        \cline{2-6}\cline{8-12}\cline{14-18}
        \rule{0pt}{2.6ex}
        Model $\backslash$ Criteria & SRCC$\uparrow$ & KRCC$\uparrow$ & PLCC$\uparrow$ & RMSE$\downarrow$ & PWRC$\uparrow$ && SRCC$\uparrow$ & KRCC$\uparrow$ & PLCC$\uparrow$ & RMSE$\downarrow$ & PWRC$\uparrow$ && SRCC$\uparrow$ & KRCC$\uparrow$ & PLCC$\uparrow$ & RMSE$\downarrow$ & PWRC$\uparrow$\\
        \midrule 
        only ranking loss	&		0.8413	&	0.6549	&	0.8607	&	7.0817	&	11.799	&	&	0.7508	&	0.5550	&	0.7596	&	7.0379	&	8.9663	&	&	0.6673	&	0.4881	&	0.6950	&	6.7167	&	7.0033	\\
        w/o CA \& SA	&		0.8647	&	0.6830	&	0.8794	&	6.5994	&	12.103	&	&	0.7834	&	0.5875	&	0.7885	&	6.6417	&	9.3234	&	&	0.6973	&	0.5122	&	0.7306	&	6.3735	&	7.3048	\\
        w/o SA	&		0.8797	&	0.7015	&	0.8917	&	6.2746	&	12.270	&	&	0.8046	&	0.6094	&	0.8094	&	6.3477	&	9.5354	&	&	0.7216	&	0.5326	&	0.7458	&	6.2209	&	7.4950	\\
        w/o CA	&		0.9160	&	0.7490	&	0.9216	&	5.4075	&	12.597	&	&	0.8618	&	0.6711	&	0.8612	&	5.5065	&	9.9569	&	&	0.7925	&	0.5982	&	0.8039	&	5.5480	&	8.0077	\\
        w/o ranking loss	&		0.9069	&	0.7363	&	0.9139	&	5.6435	&	12.507	&	&	0.8473	&	0.6545	&	0.8469	&	5.7586	&	9.8297	&	&	0.7715	&	0.5746	&	0.7864	&	5.7731	&	7.8845	\\
        w/o scoring loss	&		0.9139	&	0.7466	&	0.9195	&	5.4042	&	12.611	&	&	0.8600	&	0.6697	&	0.8618	&	5.4442	&	9.9765	&	&	0.7917	&	0.5982	&	0.8076	&	5.4694	&	8.0395	\\
        all combined	&		\textbf{0.9203}	&	\textbf{0.7550}	&	\textbf{0.9258}	&	\textbf{5.2636}	&	\textbf{12.652}	&	&	\textbf{0.8681}	&	\textbf{0.6782}	&	\textbf{0.8700}	&	\textbf{5.3412}	&	\textbf{10.011}	&	&	\textbf{0.8055}	&	\textbf{0.6105}	&	\textbf{0.8168}	&	\textbf{5.3859}	&	\textbf{8.1268}	\\
        \bottomrule
        
    \end{tabular}
    \vspace{-16pt}
\end{table*}
}

\newcommand{\TabExperimentIII}{
\begin{table*}
  \vspace{-10pt}
  \renewcommand{\arraystretch}{0.85}
  \setlength{\tabcolsep}{2.5pt}
  \fontsize{6.9pt}{\baselineskip}\selectfont
  \caption{Performance of the three variants of the state-of-the-art FR-IQA models on the ARIQA database. The  top 3 results of all three variants are in \textbf{bold} for each group. The performance changes compared to Type I in terms of SRCC are indicated in \gray{gray} fonts}
  \vspace{-5pt}
  \label{tab:3_AR_base}
  \begin{tabular}{l@{\hskip 7pt} x{20}x{20}x{20}x{20}x{20} x{2.5} x{49}x{20}x{20}x{20}x{20} x{2.5} x{49}x{20}x{20}x{20}x{20}}
    \toprule
    Method & \multicolumn{5}{c}{Type I} && \multicolumn{5}{c}{Type II} && \multicolumn{5}{c}{Type III} \\
    \cline{2-6}\cline{8-12}\cline{14-18}
    \rule{0pt}{2.6ex}
    Model $\backslash$ Criteria & SRCC$\uparrow$ & KRCC$\uparrow$ & PLCC$\uparrow$ & RMSE$\downarrow$ & PWRC$\uparrow$ && SRCC$\uparrow$ & KRCC$\uparrow$ & PLCC$\uparrow$ & RMSE$\downarrow$ & PWRC$\uparrow$ && SRCC$\uparrow$ & KRCC$\uparrow$ & PLCC$\uparrow$ & RMSE$\downarrow$ & PWRC$\uparrow$\\
    \midrule 
    PSNR	&	0.2197	&	0.1485	&	0.2742	&	12.733	&	2.8868	&	&	0.0064	\gray{(\textminus 0.2133)}	&	0.0027	&	0.0592	&	13.217	&	0.0299	&	&	0.3809	\gray{(+0.1612)}	&	0.2662	&	0.4154	&	11.901	&	4.8525	\\
    NQM \cite{damera2000image}	&	0.4101	&	0.2813	&	0.4268	&	11.974	&	5.9685	&	&	0.5348	\gray{(+0.1247)}	&	0.3772	&	0.5550	&	11.014	&	7.7948	&	&	0.5588	\gray{(+0.1487)}	&	0.4031	&	0.5867	&	10.677	&	7.6133	\\
    MS-SSIM \cite{wang2003multiscale}	&	0.6118	&	0.4414	&	0.6483	&	10.080	&	8.3464	&	&	0.6557	\gray{(+0.0439)}	&	0.4778	&	0.6609	&	9.9366	&	9.0159	&	&	0.6660	\gray{(+0.0541)}	&	0.4914	&	0.6741	&	9.6721	&	9.0949	\\
    SSIM \cite{wang2004image}	&	0.5327	&	0.3799	&	0.5551	&	11.013	&	7.2294	&	&	0.5399	\gray{(+0.0072)}	&	0.3797	&	0.5411	&	11.134	&	7.5044	&	&	0.6090	\gray{(+0.0763)}	&	0.4430	&	0.6233	&	10.276	&	8.1620	\\
    IFC \cite{sheikh2005information}	&	0.3539	&	0.2456	&	0.3294	&	12.501	&	5.6002	&	&	0.5121	\gray{(+0.1582)}	&	0.3523	&	0.5105	&	11.385	&	7.2188	&	&	0.5090	\gray{(+0.1551)}	&	0.3601	&	0.5217	&	11.172	&	7.0657	\\
    VIF \cite{sheikh2006image}	&	0.5981	&	0.4273	&	0.6366	&	10.211	&	8.4390	&	&	0.6927	\gray{(+0.0946)}	&	0.5009	&	0.6869	&	9.6218	&	9.5307	&	&	\textbf{0.7227}	\gray{(+0.1245)}	&	\textbf{0.5351}	&	\textbf{0.7222}	&	\textbf{9.2024}	&	\textbf{9.8505}	\\
    IW-MSE \cite{wang2010information}	&	0.2287	&	0.1555	&	0.2966	&	12.644	&	3.0131	&	&	0.2406	\gray{(+0.0119)}	&	0.1689	&	0.2956	&	12.648	&	3.7931	&	&	0.4126	\gray{(+0.1839)}	&	0.2906	&	0.4586	&	11.693	&	5.7220	\\
    IW-PSNR \cite{wang2010information}	&	0.2287	&	0.1555	&	0.2998	&	12.631	&	2.8814	&	&	0.2406	\gray{(+0.0119)}	&	0.1689	&	0.2895	&	12.673	&	3.7414	&	&	0.3559	\gray{(+0.1272)}	&	0.2574	&	0.4151	&	11.879	&	4.9256	\\
    IW-SSIM \cite{wang2010information}	&	0.6431	&	0.4663	&	0.6532	&	10.026	&	8.8683	&	&	\textbf{0.7103}	\gray{(+0.0672)}	&	\textbf{0.5267}	&	\textbf{0.7100}	&	\textbf{9.3231}	&	\textbf{9.7540}	&	&	\textbf{0.7116}	\gray{(+0.0685)}	&	\textbf{0.5337}	&	\textbf{0.7201}	&	\textbf{9.0193}	&	\textbf{9.6804}	\\
    FSIM \cite{zhang2011fsim}	&	0.6323	&	0.4546	&	0.6723	&	9.8010	&	8.7569	&	&	0.6538	\gray{(+0.0215)}	&	0.4716	&	0.6528	&	10.029	&	9.0221	&	&	0.6663	\gray{(+0.0340)}	&	0.4865	&	0.6774	&	9.5764	&	9.3018	\\
    GSI \cite{liu2011image}	&	0.4393	&	0.3046	&	0.5034	&	11.440	&	6.2671	&	&	0.3788	\gray{(\textminus 0.0605)}	&	0.2606	&	0.3890	&	12.197	&	5.3120	&	&	0.4245	\gray{(\textminus 0.0147)}	&	0.3056	&	0.4584	&	11.680	&	5.5771	\\
    GMSD \cite{xue2013gradient}	&	0.6485	&	0.4718	&	0.6759	&	9.7575	&	8.8107	&	&	0.5947	\gray{(\textminus 0.0537)}	&	0.4346	&	0.5959	&	10.633	&	7.7722	&	&	0.6730	\gray{(+0.0245)}	&	0.4973	&	0.6801	&	9.5815	&	9.2104	\\
    GMSM \cite{xue2013gradient}	&	0.6386	&	0.4628	&	0.6907	&	9.5745	&	8.7291	&	&	0.5863	\gray{(\textminus 0.0523)}	&	0.4142	&	0.5923	&	10.667	&	8.0777	&	&	0.6294	\gray{(\textminus 0.0092)}	&	0.4587	&	0.6422	&	10.064	&	8.5228	\\
    PAMSE \cite{xue2013perceptual}	&	0.2162	&	0.1458	&	0.2736	&	12.735	&	2.8281	&	&	0.0090	\gray{(\textminus 0.2072)}	&	0.0048	&	0.0659	&	13.211	&	0.2350	&	&	0.3657	\gray{(+0.1495)}	&	0.2558	&	0.4093	&	11.941	&	4.9560	\\
    LTG \cite{gu2014efficient}	&	0.6592	&	0.4830	&	0.6826	&	9.6759	&	8.9866	&	&	0.6469	\gray{(\textminus 0.0123)}	&	0.4742	&	0.6422	&	10.150	&	8.7329	&	&	0.6764	\gray{(+0.0172)}	&	0.4998	&	0.6818	&	9.4727	&	9.3129	\\
    VSI \cite{zhang2014vsi}	&	0.5190	&	0.3691	&	0.5926	&	10.665	&	7.3012	&	&	0.6096	\gray{(+0.0906)}	&	0.4318	&	0.6167	&	10.422	&	8.6713	&	&	0.6321	\gray{(+0.1131)}	&	0.4590	&	0.6484	&	10.039	&	8.4581	\\
    \midrule																																			
    LPIPS (Squeeze) \cite{zhang2018unreasonable}	&	0.5924	&	0.4326	&	0.6160	&	10.430	&	7.8301	&	&	0.6260	\gray{(+0.0336)}	&	0.4450	&	0.6251	&	10.334	&	8.7742	&	&	0.6417	\gray{(+0.0494)}	&	0.4693	&	0.6660	&	9.8086	&	8.7256	\\
    LPIPS (Alex) \cite{zhang2018unreasonable}	&	0.5870	&	0.4273	&	0.6314	&	10.267	&	7.7542	&	&	0.6306	\gray{(+0.0436)}	&	0.4457	&	0.6352	&	10.226	&	8.8424	&	&	0.6626	\gray{(+0.0757)}	&	0.4820	&	0.6767	&	9.6071	&	8.9045	\\
    LPIPS (VGG) \cite{zhang2018unreasonable}	&	0.5436	&	0.3828	&	0.5593	&	10.975	&	7.5461	&	&	0.6202	\gray{(+0.0766)}	&	0.4426	&	0.6141	&	10.450	&	8.6116	&	&	0.6373	\gray{(+0.0936)}	&	0.4606	&	0.6475	&	9.9848	&	8.4863	\\
    DISTS \cite{ding2020image}	&	0.5011	&	0.3583	&	0.5280	&	11.244	&	7.4008	&	&	0.5112	\gray{(+0.0101)}	&	0.3627	&	0.5528	&	11.033	&	6.6245	&	&	0.6334	\gray{(+0.1323)}	&	0.4608	&	0.6580	&	9.7866	&	8.8983	\\
    Baseline (SqueezeNet)	&	0.5733	&	0.4166	&	0.6096	&	10.496	&	7.5174	&	&	0.6339	\gray{(+0.0606)}	&	0.4570	&	0.6358	&	10.220	&	8.8444	&	&	0.6272	\gray{(+0.0539)}	&	0.4573	&	0.6493	&	9.8747	&	8.7602	\\
    Baseline (AlexNet)	&	0.5273	&	0.3776	&	0.5814	&	10.772	&	6.9367	&	&	0.6450	\gray{(+0.1177)}	&	0.4690	&	0.6578	&	9.9728	&	9.0323	&	&	0.6460	\gray{(+0.1187)}	&	0.4768	&	0.6707	&	9.7499	&	8.9168	\\
    Baseline (VGG-16)	&	0.5541	&	0.3908	&	0.5706	&	10.873	&	7.7488	&	&	0.6368	\gray{(+0.0827)}	&	0.4585	&	0.6372	&	10.204	&	8.8709	&	&	0.6587	\gray{(+0.1046)}	&	0.4805	&	0.6622	&	9.8906	&	8.8234	\\
    Baseline+ (VGG-16)	&	0.6167	&	0.4454	&	0.5649	&	10.925	&	8.3908	&	&	0.6793	\gray{(+0.0626)}	&	0.4979	&	0.6814	&	9.6902	&	9.4179	&	&	0.6815	\gray{(+0.0649)}	&	0.5036	&	0.6896	&	9.4713	&	9.4016	\\
    Baseline (VGG-19)	&	0.5612	&	0.3981	&	0.5790	&	10.795	&	7.8228	&	&	0.6561	\gray{(+0.0949)}	&	0.4750	&	0.6530	&	10.028	&	9.1239	&	&	0.6613	\gray{(+0.1001)}	&	0.4838	&	0.6674	&	9.6720	&	9.1233	\\
    Baseline (ResNet-18)	&	0.5438	&	0.3892	&	0.5750	&	10.832	&	7.4822	&	&	0.6467	\gray{(+0.1029)}	&	0.4678	&	0.6451	&	10.117	&	9.0764	&	&	0.6485	\gray{(+0.1047)}	&	0.4779	&	0.6702	&	9.7504	&	9.0147	\\
    Baseline (ResNet-34)	&	0.5426	&	0.3862	&	0.5771	&	10.813	&	7.4832	&	&	0.6603	\gray{(+0.1177)}	&	0.4782	&	0.6660	&	9.8765	&	9.2702	&	&	0.6710	\gray{(+0.1284)}	&	0.4959	&	0.6903	&	9.4826	&	9.1111	\\
    Baseline (ResNet-50)	&	0.5753	&	0.4113	&	0.5977	&	10.615	&	7.9505	&	&	0.6510	\gray{(+0.0757)}	&	0.4688	&	0.6464	&	10.102	&	9.1610	&	&	0.6619	\gray{(+0.0866)}	&	0.4831	&	0.6736	&	9.7163	&	8.9478	\\
    
    CFIQA (VGG-19)	&	0.5002	&	0.3588	&	0.4881	&	11.556	&	6.5740	&	&	\textbf{0.7448}	\gray{(+0.2446)}	&	\textbf{0.5539}	&	\textbf{0.7436}	&	\textbf{8.8522}	&	\textbf{10.212}	&	&	\textbf{0.7511}	\gray{(+0.2509)}	&	\textbf{0.5648}	&	\textbf{0.7587}	&	\textbf{8.4569}	&	\textbf{10.080}	\\
    CFIQA (ResNet-34)	&	0.4061	&	0.2903	&	0.4405	&	11.886	&	5.3775	&	&	0.7024	\gray{(+0.2963)}	&	0.5178	&	0.7092	&	9.3344	&	\textbf{9.6762}	&	&	\textbf{0.7092}	\gray{(+0.3031)}	&	\textbf{0.5280}	&	\textbf{0.7197}	&	\textbf{9.0977}	&	9.6462	\\

    \bottomrule
  \end{tabular}
  \vspace{-15pt}
\end{table*}
}

\newcommand{\TabExperimentIV}{
\begin{table}
\vspace{-1pt}
  \renewcommand{\arraystretch}{0.85}
  \setlength{\tabcolsep}{4.6pt}
  \fontsize{7.2pt}{\baselineskip}\selectfont
  \caption{Performance of four trainable models.}
  \vspace{-5pt}
  \label{tab:4_ARIQA}
  \centering
  \begin{tabular}{cccccc}
    \toprule
    Model $\backslash$ Criteria & SRCC$\uparrow$ & KRCC$\uparrow$ & PLCC$\uparrow$ & RMSE$\downarrow$ & PWRC$\uparrow$\\
    \midrule 
    LPIPS	&	0.7624	&	0.5756	&	0.7591	&	8.6935	&	10.936 	\\
    CFIQA	&	0.7787	&	0.5863	&	0.7695	&	8.5484	&	11.125 	\\
    ARIQA	&	0.7902	&	0.5967	&	0.7824	&	8.3295	&	11.314	\\
    ARIQA+	&	\textbf{0.8124}	&	\textbf{0.6184}	&	\textbf{0.8136}	&	\textbf{7.8018} & \textbf{11.576}	\\
    \bottomrule
  \end{tabular}
  \vspace{-15pt}
\end{table}
}